\documentclass{ieeeaccess}
\usepackage{cite}
\usepackage{amsmath,amssymb,amsfonts}
\usepackage{mathtools}
\usepackage{algorithmic}
\usepackage{graphicx}
\usepackage{textcomp}
\usepackage{wasysym}
\usepackage{booktabs}
\usepackage[caption=false, font=footnotesize]{subfig}

\usepackage{epstopdf}
\usepackage{units}
\usepackage{pgfplots}
\usepackage{tikz}
\usepgfplotslibrary{external} 
\pgfplotsset{compat=1.14,every tick label/.append style={font=\footnotesize}}
\tikzset{every picture/.style={font issue=\footnotesize},
	font issue/.style={execute at begin picture={#1\selectfont}}
}
\usetikzlibrary{external,arrows.meta,patterns,ipe} 

\usepackage{hyperref}
\usepackage{lipsum}

\usepackage{dcolumn}
\newcolumntype{L}{D{.}{.}{2,4}}

\NewSpotColorSpace{PANTONE}
\AddSpotColor{PANTONE} {PANTONE3015C} {PANTONE\SpotSpace 3015\SpotSpace C} {1 0.3 0 0.2}
\SetPageColorSpace{PANTONE}%

\def\BibTeX{{\rm B\kern-.05em{\sc i\kern-.025em b}\kern-.08em
    T\kern-.1667em\lower.7ex\hbox{E}\kern-.125emX}}

\newcommand\copyrighttext{%
	\footnotesize \textcopyright 2019 IEEE.  Personal use of this material is permitted.  Permission from IEEE must be obtained for all other uses, in any current or future media, including reprinting/republishing this material for advertising or promotional purposes, creating new collective works, for resale or redistribution to servers or lists, or reuse of any copyrighted component of this work in other works.}
\newcommand\copyrightnotice{%
	\begin{tikzpicture}[remember picture,overlay]
	\node[anchor=north,yshift=-10pt] at (current page.north) {\fbox{\parbox{\dimexpr\textwidth-\fboxsep-\fboxrule\relax}{\copyrighttext}}};
	\end{tikzpicture}%
}

\begin{document}
\history{Date of publication xxxx 00, 0000, date of current version xxxx 00, 0000.}
\doi{10.1109/ACCESS.2019.2956882}

\title{Ground truth force distribution for learning-based tactile sensing:\newline a finite element approach}
\author{\uppercase{Carmelo Sferrazza}\authorrefmark{1}, \IEEEmembership{Student Member, IEEE},
\uppercase{Adam Wahlsten}\authorrefmark{2}, \uppercase{Camill Trueeb}\authorrefmark{1}, \uppercase{and Raffaello D'Andrea}\authorrefmark{1}, \IEEEmembership{Fellow, IEEE}}
\address[1]{Institute for Dynamic Systems and Control, ETH Zurich, 8092 Zurich, Switzerland}
\address[2]{Institute for Mechanical Systems, ETH Zurich, 8092 Zurich, Switzerland}
\tfootnote{AW was supported by funding from the Swiss National Science Foundation (Grant No. 179012).}

\markboth
{C. Sferrazza \headeretal: Ground truth force distribution for learning-based tactile sensing: a finite element approach}
{C. Sferrazza \headeretal: Ground truth force distribution for learning-based tactile sensing: a finite element approach}

\corresp{Corresponding author: Carmelo Sferrazza (e-mail: csferrazza@ethz.ch).}
\begin{abstract}
Skin-like tactile sensors provide robots with rich feedback related to the force distribution applied to their soft surface. The complexity of interpreting raw tactile information has driven the use of machine learning algorithms to convert the sensory feedback to the quantities of interest. However, the lack of ground truth sources for the entire contact force distribution has mainly limited these techniques to the sole estimation of the total contact force and the contact center on the sensor's surface. The method presented in this article uses a finite element model to obtain ground truth data for the three-dimensional force distribution. The model is obtained with state-of-the-art material characterization methods and is evaluated in an indentation setup, where it shows high agreement with the measurements retrieved from a commercial force-torque sensor. The proposed technique is applied to a vision-based tactile sensor, which aims to reconstruct the contact force distribution purely from images. Thousands of images are matched to ground truth data and are used to train a neural network architecture, which is suitable for real-time predictions.
\end{abstract}


\titlepgskip=-15pt

\maketitle
\thispagestyle{empty}
\tikzexternaldisable 
\copyrightnotice
\tikzexternalenable

\vspace{-15pt}

\section{Introduction} \label{sec:introduction}
A growing number of applications require robots to interact with the environment \cite{bin_picking} and with humans \cite{robots_at_home}. The use of soft materials for robotics applications \cite{nature_news} introduces intrinsic safety during interactive tasks \cite{intrinsic_safety}. In addition, precise estimation of contact forces is crucial for effective operation without damaging the robot's surroundings, e.g., for manipulation of fragile objects \cite{fragile_object}.

Modeling the interaction of soft materials with generic objects is highly complex. As a consequence, several tactile sensing strategies leverage the use of machine learning algorithms to map sensory feedback to the corresponding quantities of interest, e.g., contact forces, shape and materials, see \cite{contact_materials,contact_quantities,vision_based_training}. These maps are generally retrieved by means of supervised learning techniques, which fit a model to a large amount of labeled data, i.e., sensory data paired with the corresponding ground truth.

However, the estimation of the full contact force distribution purely from data is limited by the lack of a ground truth source that does not alter the interaction between the soft material and the objects in contact. This article aims to provide a systematic way of labeling data with ground truth for the three-dimensional force distribution, which is obtained in simulation through the finite element method (FEM). 

The approach is evaluated on a vision-based tactile sensor, originally presented in \cite{sferrazza_sensors}, which uses a camera to track spherical particles within a transparent gel. Hyperelastic models of the sensor's materials are retrieved from state-of-the-art material characterization tests, which are fully independent of the evaluation experiments. A label vector representing the ground truth force distribution is assigned to each image collected during an automatic indentation procedure. The total contact force also retrieved from the FEM simulations shows a satisfactory agreement with the measurements obtained from a commercial force-torque (F/T) sensor.

\Figure[h!](topskip=0pt, botskip=0pt, midskip=0pt)[width=0.99\columnwidth]{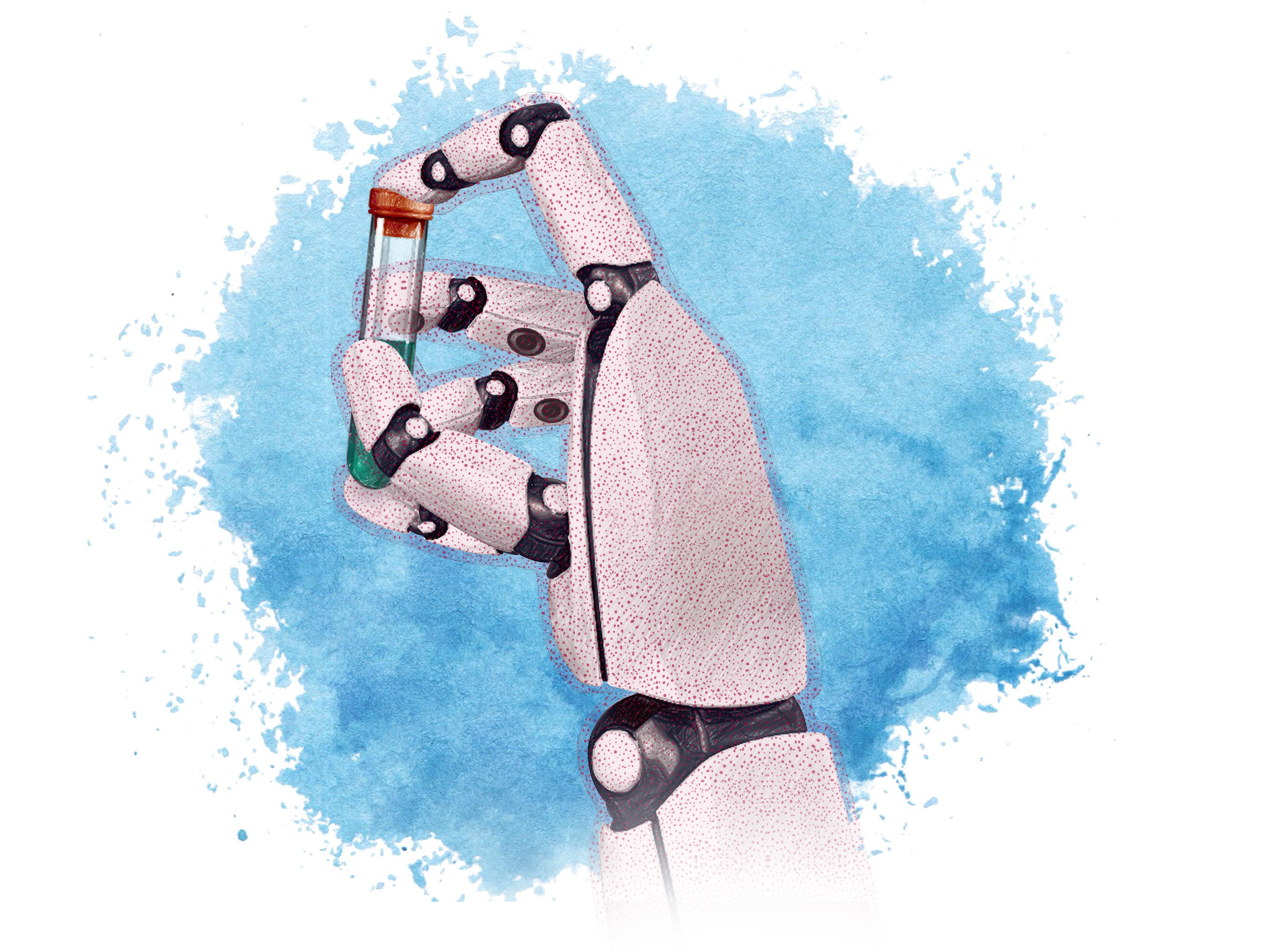}
{The tactile sensing technique presented in \cite{sferrazza_sensors} is suitable to cover large surfaces of arbitrary shape and dimension. A concept of a robotic hand covered with cameras and a transparent gel embedding a spread of red particles is shown in this figure.\label{fig:concept}}

The dataset generated with the strategy proposed here is then used to train a deep neural network (DNN) architecture \cite{deep_learning_book}, which maps optical flow features to the contact force distribution. The evaluation of this strategy is carried out in a specific indentation setup, with the resulting pipeline running in real-time on the CPU of a standard laptop computer (dual-core, 2.80 GHz) at 40 Hz.

\subsection{Related work}
In recent decades, tactile sensing research has shown the potential of providing robots with the sense of touch, exploited both singly \cite{blind_grasping} or in combination with vision \cite{more_than_a_feeling}.

Among the various categories of tactile sensors, see \cite{survey_ts,survey_ts2} for a survey, vision-based (or optical) tactile sensors are based on optical devices that monitor properties related to the contact between the sensor's surface (generally soft) and the environment. Among the advantages of this type of tactile sensors are high resolution, low cost, ease of manufacture and the preservation of the surface softness.

One category of optical tactile sensors uses a camera to track sparse markers within a soft, transparent gel, which deforms when subject to external forces, see for example \cite{fingervision,tactip}. Other optical devices are able to provide information about the contact with the environment, as shown with the use of dynamic vision sensors \cite{dvs_ts} and depth cameras \cite{soft_bubble}. The sensor used here for the evaluation of the proposed approach is based on an RGB camera (which retains a small size) that tracks a dense spread of particles within a soft gel, see for example Fig. \ref{fig:concept}. This design is presented in \cite{sferrazza_sensors} and shows performance advantages over sparse marker tracking, and ease of manufacture, without any assumptions about the surface shape.

Converting tactile information to quantities, such as the force distribution, which are of high relevance for many robotics tasks (e.g., grasping or manipulation), is not trivial. In fact, the main complexity is introduced by the absence of a generally valid closed-form model, which maps the deformation of a hyperelastic material to the external forces applied to it. The use of data-driven techniques aims to overcome this problem, approximating this map with a model learned from a collection of past data. In \cite{gelsight_sensors}, an optical tactile sensor that exploits photometric stereo and markers painted on its soft surface is used to reconstruct the total contact force by means of a neural network architecture. In \cite{overlapping_signals}, an array of light emitters and receivers is placed below a soft gel to create tactile information, which is then provided to machine learning algorithms that reconstruct the location and the depth of an indentation, as well as the type of the employed indenter. Although these techniques generally require large datasets, transfer learning techniques can reuse information extracted across different sensors, as shown in \cite{transfer_learning_sferrazza}.

The FEM \cite{hutton_fem} is a powerful numerical technique that provides approximate solutions of boundary value problems arising in engineering, by subdividing a large system into many smaller parts (called elements). One of the widespread applications of this technique is the analysis of the behavior of soft materials under various loading conditions. In \cite{inverse_fem}, the silicone gel pad of an optical tactile sensor is modeled as a linear elastic material, and the FEM is used to compute the stiffness matrix that approximates the relation between external forces and displacements of the sensor's material. Based on reconstructed surface displacements, this matrix is then used to compute an estimate of the force distribution applied to the sensor. FEM simulations of a flexible 3D shape sensor are used in \cite{shape_fea} to optimize design parameters. Furthermore, these simulations show the uniqueness of a strain-to-shape mapping for the case considered. In \cite{realtime_fem}, theoretical justifications based on FEM equations are provided to reconstruct the external forces applied to soft bodies using information about their deformation.

The strategy followed in this article exploits FEM simulations to obtain ground truth data for the contact force distribution applied to the soft surface of a tactile sensor. The lack of ground truth data has so far prevented the development of learning-based tactile sensors that predict the full force distribution, limiting them to the estimation of simpler quantities, e.g., the resultant force and its location, and the depth of a contact. The hyperelastic models identified capture the full material behavior, including nonlinearities, rendering highly accurate simulations. Images collected in experiments on a vision-based tactile sensor are matched to the ground truth and used to train a DNN that reconstructs the force distribution with high accuracy and in real-time. 

\subsection{Outline}
The sensing principle and the hardware used for the evaluation are presented in Section \ref{sec:sensing_principle}, while the material characterization is discussed in Section \ref{sec:material_characterization}. In Section \ref{sec:data_collection}, the dataset generation is described, from the collection of images to the approach proposed for assigning ground truth labels. The learning algorithm and the results are presented in Section \ref{sec:learning}. Section \ref{sec:conclusion} concludes the article with a brief discussion.

\section{Hardware} \label{sec:sensing_principle}
\begin{figure}
	\centering
	\subfloat[Sensor protoype]{%
		\includegraphics[width=0.48\linewidth]{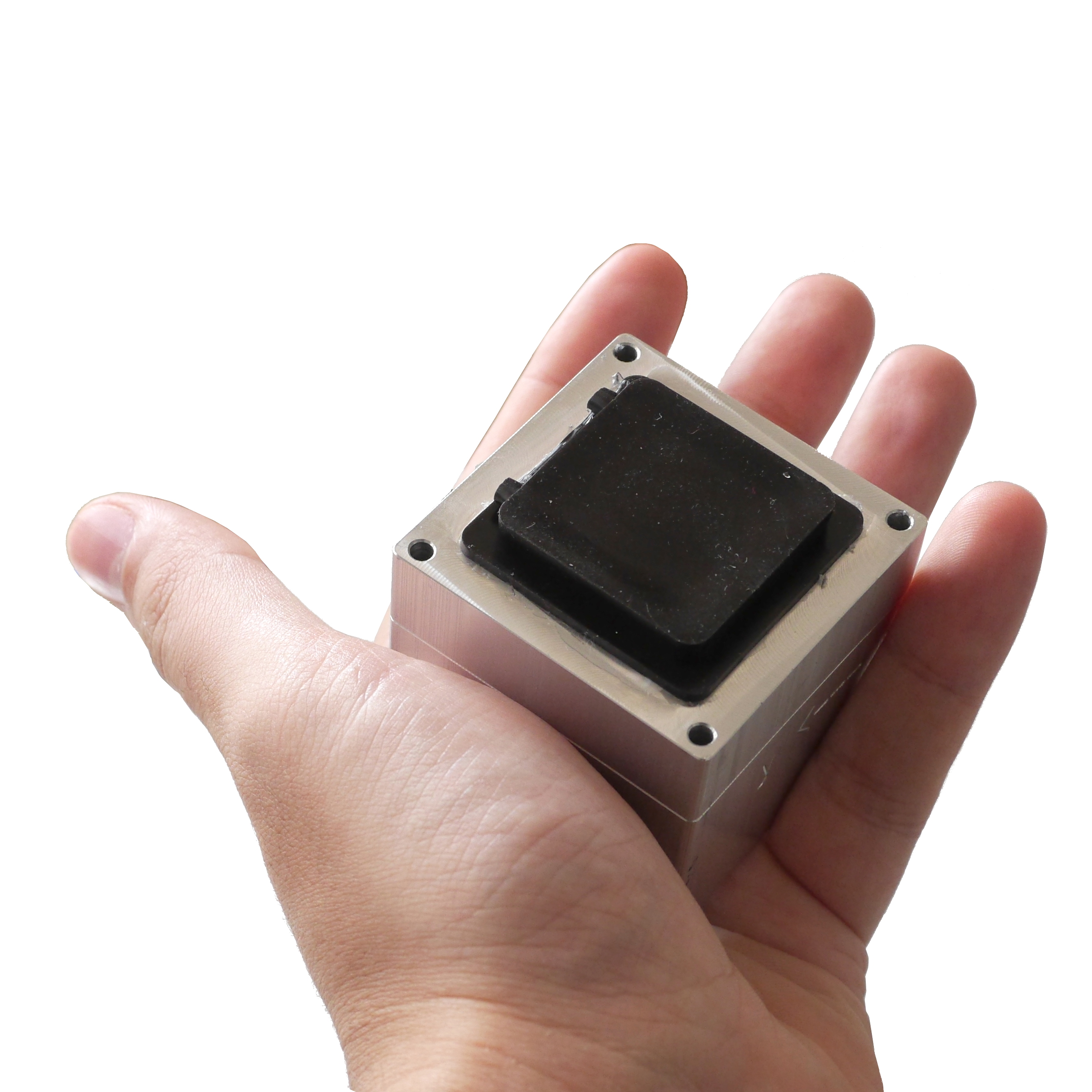} }
	\hfill
	\subfloat[Particles image]{%
		\includegraphics[width=0.48\linewidth]{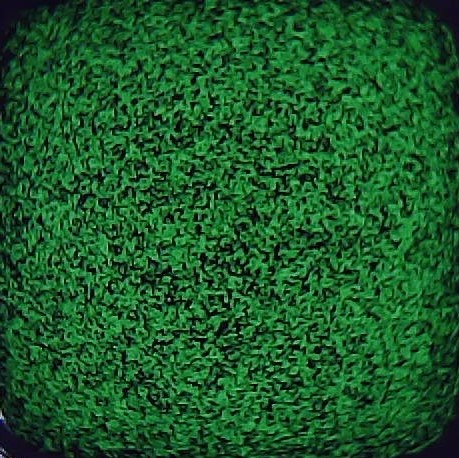} }
	\caption{The prototype of the sensor (designed for desktop testing) is shown in (a). The dense spread of green particles is captured by the camera placed inside the aluminum part. The resulting RGB image at a state of zero force is shown in (b).}
	\label{fig:hardware}
\end{figure}

The approach discussed in this article for generating ground truth labels is evaluated on a vision-based tactile sensor. The tactile sensing strategy is presented in \cite{sferrazza_sensors}, and is based on tracking the movement of spherical particles, which are randomly spread within a soft gel placed in front of a camera. The prototype used for the experiments and the camera image at a state of zero force are shown in Fig. \ref{fig:hardware}.

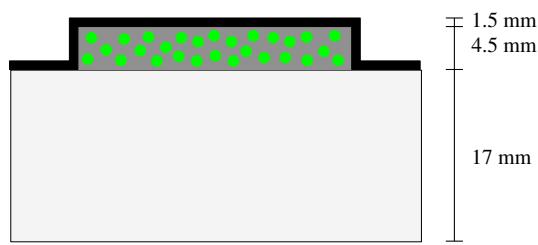
\begin{figure}
	\centering
	\scalebox{0.2}{\tikzstyle{ipe stylesheet} = [
  ipe import,
  even odd rule,
  line join=round,
  line cap=butt,
  ipe pen normal/.style={line width=0.4},
  ipe pen heavier/.style={line width=0.8},
  ipe pen fat/.style={line width=1.2},
  ipe pen ultrafat/.style={line width=2},
  ipe pen normal,
  ipe mark normal/.style={ipe mark scale=3},
  ipe mark large/.style={ipe mark scale=5},
  ipe mark small/.style={ipe mark scale=2},
  ipe mark tiny/.style={ipe mark scale=1.1},
  ipe mark normal,
  /pgf/arrow keys/.cd,
  ipe arrow normal/.style={scale=7},
  ipe arrow large/.style={scale=10},
  ipe arrow small/.style={scale=5},
  ipe arrow tiny/.style={scale=3},
  ipe arrow normal,
  /tikz/.cd,
  ipe arrows, 
  <->/.tip = ipe normal,
  ipe dash normal/.style={dash pattern=},
  ipe dash dashed/.style={dash pattern=on 4bp off 4bp},
  ipe dash dotted/.style={dash pattern=on 1bp off 3bp},
  ipe dash dash dotted/.style={dash pattern=on 4bp off 2bp on 1bp off 2bp},
  ipe dash dash dot dotted/.style={dash pattern=on 4bp off 2bp on 1bp off 2bp on 1bp off 2bp},
  ipe dash normal,
  ipe node/.append style={font=\normalsize},
  ipe stretch normal/.style={ipe node stretch=1},
  ipe stretch normal,
  ipe opacity 10/.style={opacity=0.1},
  ipe opacity 30/.style={opacity=0.3},
  ipe opacity 50/.style={opacity=0.5},
  ipe opacity 75/.style={opacity=0.75},
  ipe opacity opaque/.style={opacity=1},
  ipe opacity opaque,
]
\definecolor{red}{rgb}{1,0,0}
\definecolor{green}{rgb}{0,1,0}
\definecolor{blue}{rgb}{0,0,1}
\definecolor{yellow}{rgb}{1,1,0}
\definecolor{orange}{rgb}{1,0.647,0}
\definecolor{gold}{rgb}{1,0.843,0}
\definecolor{purple}{rgb}{0.627,0.125,0.941}
\definecolor{gray}{rgb}{0.745,0.745,0.745}
\definecolor{brown}{rgb}{0.647,0.165,0.165}
\definecolor{navy}{rgb}{0,0,0.502}
\definecolor{pink}{rgb}{1,0.753,0.796}
\definecolor{seagreen}{rgb}{0.18,0.545,0.341}
\definecolor{turquoise}{rgb}{0.251,0.878,0.816}
\definecolor{violet}{rgb}{0.933,0.51,0.933}
\definecolor{darkblue}{rgb}{0,0,0.545}
\definecolor{darkcyan}{rgb}{0,0.545,0.545}
\definecolor{darkgray}{rgb}{0.663,0.663,0.663}
\definecolor{darkgreen}{rgb}{0,0.392,0}
\definecolor{darkmagenta}{rgb}{0.545,0,0.545}
\definecolor{darkorange}{rgb}{1,0.549,0}
\definecolor{darkred}{rgb}{0.545,0,0}
\definecolor{lightblue}{rgb}{0.678,0.847,0.902}
\definecolor{lightcyan}{rgb}{0.878,1,1}
\definecolor{lightgray}{rgb}{0.827,0.827,0.827}
\definecolor{lightgreen}{rgb}{0.565,0.933,0.565}
\definecolor{lightyellow}{rgb}{1,1,0.878}
\definecolor{black}{rgb}{0,0,0}
\definecolor{white}{rgb}{1,1,1}
\definecolor{verylightgray}{rgb}{0.957,0.957,0.957}
\definecolor{verydarkgray}{rgb}{0.563,0.563,0.563}
\begin{tikzpicture}[ipe stylesheet]
  \filldraw[ipe pen fat, fill=verylightgray]
    (34.4019, 638.7991) rectangle (802.4019, 318.7991);
  \fill[verydarkgray]
    (160, 640) rectangle (672, 720);
  \filldraw[black, ipe pen fat]
    (32, 640)
     -- (32, 656)
     -- (144, 656)
     -- (144, 736)
     -- (688, 736)
     -- (688, 656)
     -- (800, 656)
     -- (800, 640)
     -- (672, 640)
     -- (672, 720)
     -- (160, 720)
     -- (160, 640)
     -- (32, 640);
  \filldraw[green, ipe pen fat]
    (185.457, 699.033) circle[radius=9.9032];
  \filldraw[green, ipe pen fat]
    (213.078, 676.215) circle[radius=9.9032];
  \filldraw[green, ipe pen fat]
    (276.728, 675.014) circle[radius=9.9032];
  \filldraw[green, ipe pen fat]
    (246.7046, 696.631) circle[radius=9.9032];
  \filldraw[green, ipe pen fat]
    (353.5878, 699.033) circle[radius=9.9032];
  \filldraw[green, ipe pen fat]
    (348.7841, 664.206) circle[radius=9.9032];
  \filldraw[green, ipe pen fat]
    (384.8121, 691.827) circle[radius=9.9032];
  \filldraw[green, ipe pen fat]
    (417.2373, 665.407) circle[radius=9.9032];
  \filldraw[green, ipe pen fat]
    (447.261, 691.827) circle[radius=9.9032];
  \filldraw[green, ipe pen fat]
    (450.864, 655.799) circle[radius=9.9032];
  \filldraw[green, ipe pen fat]
    (488.092, 702.635) circle[radius=9.9032];
  \filldraw[green, ipe pen fat]
    (508.508, 661.804) circle[radius=9.9032];
  \filldraw[green, ipe pen fat]
    (525.321, 699.033) circle[radius=9.9032];
  \filldraw[green, ipe pen fat]
    (588.971, 657) circle[radius=9.9032];
  \filldraw[green, ipe pen fat]
    (620.195, 673.813) circle[radius=9.9032];
  \filldraw[green, ipe pen fat]
    (556.546, 691.827) circle[radius=9.9032];
  \filldraw[green, ipe pen fat]
    (548.139, 661.804) circle[radius=9.9032];
  \filldraw[green, ipe pen fat]
    (646.616, 657) circle[radius=9.9032];
  \filldraw[green, ipe pen fat]
    (292.3401, 700.234) circle[radius=9.9032];
  \filldraw[green, ipe pen fat]
    (178.251, 659.402) circle[radius=9.9032];
  \filldraw[green, ipe pen fat]
    (382.4102, 655.799) circle[radius=9.9032];
  \filldraw[green, ipe pen fat]
    (474.882, 673.813) circle[radius=9.9032];
  \filldraw[green, ipe pen fat]
    (414.8355, 702.635) circle[radius=9.9032];
  \filldraw[green, ipe pen fat]
    (324.7654, 681.019) circle[radius=9.9032];
  \filldraw[green, ipe pen fat]
    (307.9523, 655.799) circle[radius=9.9032];
  \filldraw[green, ipe pen fat]
    (240.7, 657) circle[radius=9.9032];
  \filldraw[green, ipe pen fat]
    (587.77, 700.234) circle[radius=9.9032];
  \filldraw[green, ipe pen fat]
    (640.611, 702.636) circle[radius=9.9032];
  \draw[ipe pen ultrafat]
    (864, 640)
     -- (864, 320);
  \draw[ipe pen ultrafat]
    (848, 640)
     -- (880, 640);
  \draw[ipe pen ultrafat]
    (848, 320)
     -- (880, 320);
  \draw[ipe pen ultrafat]
    (864, 640)
     -- (864, 720);
  \draw[ipe pen ultrafat]
    (848, 720)
     -- (880, 720);
  \draw[ipe pen ultrafat]
    (864, 736)
     -- (864, 720);
  \draw[ipe pen ultrafat]
    (848, 736)
     -- (880, 736);
  \node[ipe node, font={\fontsize{40}{48}\selectfont}]
     at (896, 720) {1.5 mm};
  \node[ipe node, font={\fontsize{40}{48}\selectfont}]
     at (896, 672) {4.5 mm};
  \node[ipe node, font={\fontsize{40}{48}\selectfont}]
     at (896, 464) {17 mm};
\end{tikzpicture}}
	\caption{A scheme of the three-layer structure that composes the soft material. The thickness of the different layers is shown in the figure above. This structure yields a top surface of 32x32 mm.}
	\label{fig:three_layers}
\end{figure}

The soft material is produced in a three-layer structure, as depicted in Fig. \ref{fig:three_layers}. From the bottom (which touches the camera lens) to the top surface, the following materials are used: 1) a stiffer layer (ELASTOSIL\textregistered{} RT 601 RTV-2, mixing ratio 7:1, shore hardness 45A); 2) the soft gel (Ecoflex\texttrademark{} GEL, mixing ratio 1:1, shore hardness 000-35) embedding the particles, which comprise 1.96 \% of the layer volume; 3) a black surface layer (ELASTOSIL\textregistered{} RT 601 RTV-2, mixing ratio 25:1, shore hardness 10A). After curing, the complete sensor is placed in an oven at 60 \textdegree C for 8 hours. This step has the effect of reducing the aging of the materials, which is discussed in further detail in Section \ref{sec:material_characterization}.

\section{Material characterization}\label{sec:material_characterization}
Finite element analysis (FEA) of arbitrary contact interactions with the sensor's soft surface requires material models that account for geometrical and material nonlinearities. Soft elastomers are often modeled as hyperelastic materials \cite{ogden_nonlinear_elasticity}, and finding a suitable model formulation and corresponding parameters generally necessitates experimental data from both uniaxial and biaxial stress states \cite{steinmann_fitting, hopf_pdms_characterization}. To this end, a large-strain multiaxial characterization of the two most compliant materials, the Ecoflex GEL and the Elastosil 25:1, is performed. Samples of both materials are tested in uniaxial tension (UA), pure shear (PS), and equibiaxial tension (EB) based on previously described protocols \cite{hopf_pdms_characterization}. The bottom layer of Elastosil with the mixing ratio 7:1 is considerably stiffer than the soft adjacent Ecoflex GEL, see Section \ref{sec:material_characterization_results}, and is therefore modeled as rigid in the subsequent FEA. 

\subsection{Sample preparation}
Thin material sheets of each elastomer are prepared as described in Section \ref{sec:sensing_principle}, and cast to a nominal thickness of 0.5~mm. Test pieces are cut to obtain gauge dimensions (length $\times$ width) of 40~mm~$\times$~10~mm for UA, 10~mm~$\times$~60~mm for PS, and a diameter of 30~mm for membrane inflation tests (EB). The test pieces of Ecoflex GEL used for the mechanical characterization do not contain the spherical particles, i.e. the pure material behavior is tested. An ink pattern is applied to the sample surface to facilitate optical strain analysis \cite{hopf_pdms_characterization}. After each experiment, the sample thickness $h_0$ is measured on cross-sections cut from the central region using a confocal microscope (LSM 5 Pascal, Carl Zeiss AG) with a 10$\times$ objective in brightfield mode. 

\subsection{Mechanical testing}
UA and PS tests are performed on a tensile testing set-up (MTS Systems) consisting of horizontal hydraulic actuators, 50~N force sensors, and a CCD-camera (Pike F-100B, Allied Vision Technologies GmbH) equipped with a 0.25$\times$ telecentric lens (NT55-349, Edmund Optics Ltd.) that captures top-view images of the deforming test piece. Displacement-controlled monotonic tests are performed up to a specified nominal strain (Ecoflex GEL: 200 \%; Elastosil 25:1: 100 \%) at a nominal strain rate of 0.3 \%/s. The strain-rate dependence is analyzed in an additional UA test, where the sample is loaded cyclically with strain rates increasing from 0.1~\%/s up to 10~\%/s.

An EB state of tension is realized in a pressure-controlled membrane inflation test (see \cite{hopf_pdms_characterization} for details). Briefly, a thin, circular sample is clamped on top of a hollow cylinder and inflated by means of a syringe pump (PhD Ultra, Harvard Apparatus), while a pressure sensor (LEX 1, Keller AG) measures the inflation pressure $p$. Top and side-view images are recorded with CCD cameras (GRAS-14S5C-C, Point Grey Research), and the image sequences are used for evaluating the in-plane deformation at the apex and the apex radius of curvature $r$, respectively.

All experiments are performed at room temperature and on the same day as completed curing. The mechanical properties of soft elastomers are known to change with aging \cite{hopf_pdms_characterization, placet_pdms_aging}, a process attributed to additional, thermally activated cross-linking \cite{placet_pdms_aging}. To assess the influence of aging, additional UA test pieces of the same sheets were kept at room temperature and tested several weeks after fabrication. 

\subsection{Experimental data analysis}
Since nominal strains computed from the clamp displacements are prone to errors due to sample slippage \cite{bernardi_elastomers}, the local in-plane principal stretches in the center of the test-piece, $\lambda_1$ and $\lambda_2$, are computed from the top-view image sequences using a custom optical flow-tracking algorithm \cite{hopf_pdms_characterization}. The principal stretch in thickness direction is calculated by assuming material incompressibility, i.e., $\lambda_3 = 1/(\lambda_1\lambda_2)$. In the UA and PS configurations, the Cauchy stress in loading direction is evaluated as $\sigma = F\lambda/(w_0 h_0)$, where $F$ are the measured force values, $\lambda\coloneqq\lambda_1$ is the principal stretch in loading direction, and $w_0$ is the reference width of the test piece. For inflation tests, the measured inflation pressure and apex radius of curvature can be used to approximate the equibiaxial Cauchy stress at the apex as $\sigma = pr/(2h_0\lambda_3)$, which holds for $h_0 \ll r$ \cite{ogden_model}.

\subsection{Constitutive models}\label{sec:constitutive_models}
The experimental data are used to fit the parameters of a hyperelastic, incompressible Ogden model \cite{ogden_model}, for which the strain-energy density per unit reference volume reads
\begin{equation}\label{eq:ogden_model}
W = \sum_{k=1}^{K} \frac{\mu_k}{\alpha_k}\left(\lambda_1^{\alpha_k} +  \lambda_2^{\alpha_k} + \lambda_3^{\alpha_k} - 3 \right), \; \lambda_1\lambda_2\lambda_3 = 1.
\end{equation}
The material parameters $\mu_k$, $\alpha_k$ must satisfy the constraint $\mu_k\alpha_k > 0$, $k=1,2,\ldots, K$, and can be used to calculate the corresponding Young's modulus as $E = (1+\nu)\sum_{k=1}^{K}\mu_k\alpha_k$, with $\nu=0.5$ being the Poisson's ratio of an isotropic incompressible material. The principal Cauchy stresses immediately follow from \eqref{eq:ogden_model} as (see \cite{ogden_model}, p. 571)
\begin{equation}\label{eq:ogden_principal_stresses}
\sigma_i = \sum_{k=1}^{K} \mu_k\lambda_i^{\alpha_k} - q, \quad i = 1,2,3,
\end{equation}
where $q$ is an arbitrary hydrostatic pressure arising due to the incompressibility constraint, whose value depends on the boundary conditions. By specializing \eqref{eq:ogden_principal_stresses} to the three experimentally considered load cases (see \cite{ogden_model}), the analytical formulas were used to minimize the squared error between the model and the experiments using the minimization routine \texttt{fmincon} available in MATLAB (R2018b, The MathWorks, Inc.). Ogden models of order $K = 2$ were found to provide the best description of the data for both materials compared to neo-Hookean or Mooney--Rivlin formulations; the resulting parameter sets are reported in Table \ref{tab:material_parameters}. 

\begin{table}[h!]
	\centering
	\caption{Material parameters of Ogden's model for the Ecoflex GEL and the Elastosil 25:1}
	\begin{tabular}{l L L L L}
		\toprule
		Material & \multicolumn{1}{c}{$\mu_1$ [kPa]} &  \multicolumn{1}{c}{$\alpha_1$ [-]} &  \multicolumn{1}{c}{$\mu_2$ [kPa]} &  \multicolumn{1}{c}{$\alpha_2$ [-]} \\
		\midrule
		Ecoflex GEL & 7.9652 & 1.2769 & 0.3093 & 3.5676\\
		Elastosil 25:1 & 85.1168 & 2.8991 & -0.0020 & -8.2915\\
		\bottomrule
	\end{tabular}
	\label{tab:material_parameters}
\end{table}

\subsection{Results}\label{sec:material_characterization_results}
\begin{figure*}
	\centering
	\subfloat[Mechanical behavior of the Ecoflex GEL]{%
	\includegraphics[scale=1.0]{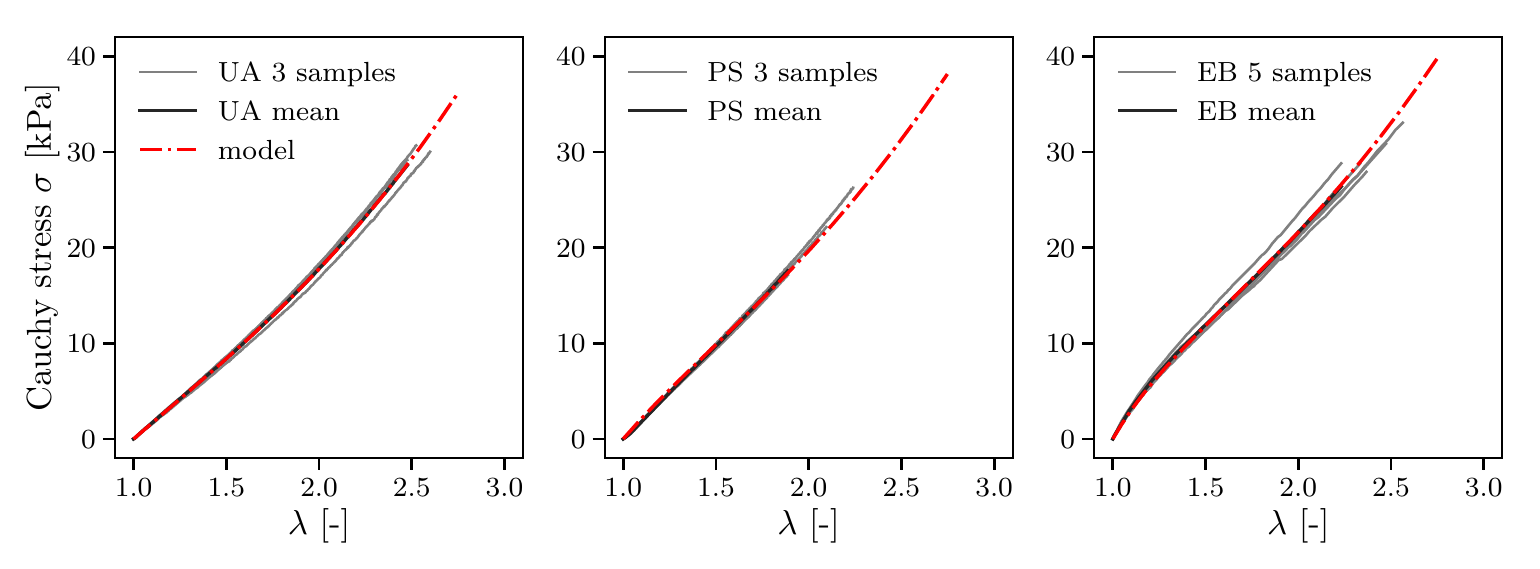}}
	\hfill
	\subfloat[Mechanical behavior of the Elastosil 25:1]{%
	\includegraphics[scale=1.0]{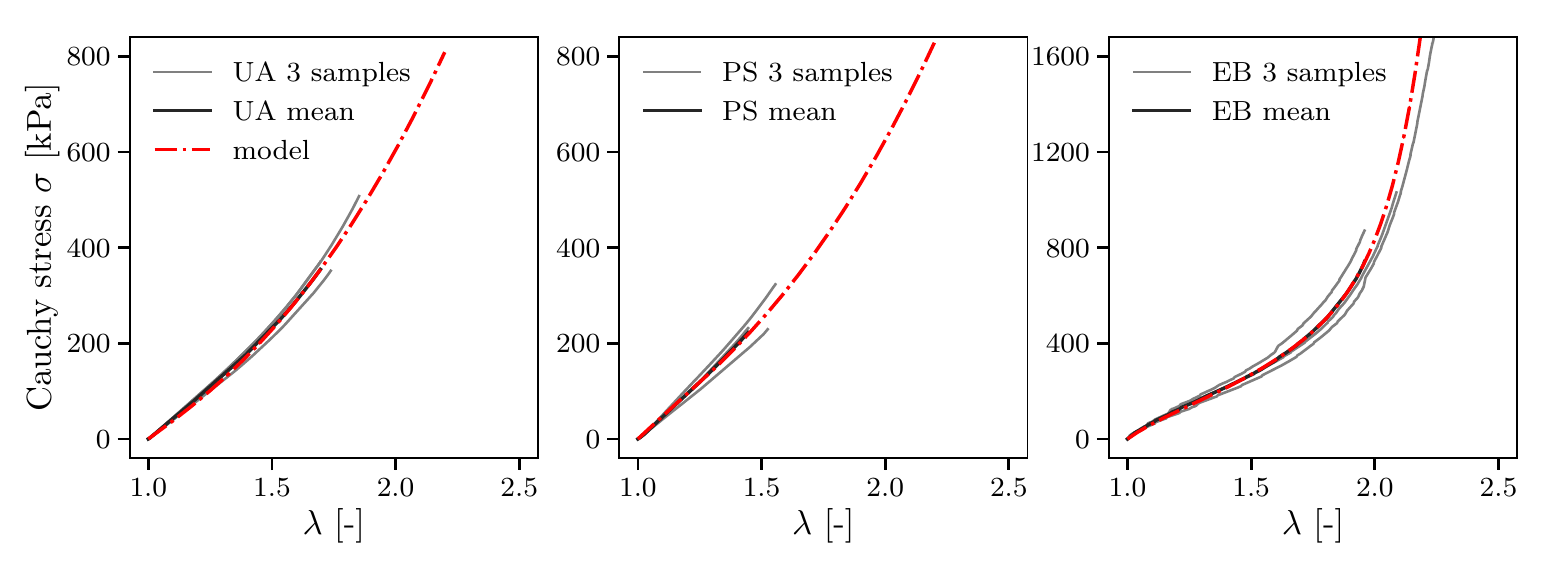}}
	\caption{Stress-stretch response of the Ecoflex GEL (a) and the Elastosil 25:1 (b) in, from left to right, uniaxial tension (UA), pure shear (PS), and equibiaxial tension (EB), together with corresponding hyperelastic model predictions. Note the different scales in (a) and (b), in particular the significantly stiffer equibiaxial response of the Elastosil 25:1.}
	\label{fig:material_characterization}
\end{figure*}

The individual stress-stretch curves for each sample of the two elastomers tested are reported in Fig. \ref{fig:material_characterization}, together with the sample averages and the model predictions. Both models identified provide an excellent description of the mechanical behavior over the whole range of deformation for all three load cases. The additional UA tests suggest a negligible influence of both strain rate and shelf time (over 5 weeks) on the mechanical behavior of the Elastosil 25:1 for the rates and times tested. However, the Ecoflex GEL shows a dependence on both strain rate and aging, see Fig. \ref{fig:rate_aging_ecoflex}, in the Appendix \ref{sec:appendix}. These dependencies, as well as potential softening phenomena upon cyclic loading (Mullins effect), are neglected in the hyperelastic model.

Corresponding Young's moduli (calculated using the Ogden model coefficients) of the Ecoflex GEL and the Elastosil 25:1 are 16.9 kPa and 370.2 kPa, respectively. For comparison, the Young's modulus of the stiffer Elastosil 7:1, determined by microindentation tests (FT-MTA02, FemtoTools AG), is found to be 0.97 MPa, i.e. more than 50 times stiffer than the Ecoflex GEL.

\section{Generating a dataset} \label{sec:data_collection}
The task of mapping the information extracted from the images to the applied contact force distribution is formulated here as a supervised learning problem. This requires a training dataset composed of input features (here retrieved from the images) and the respective ground truth labels (here obtained from finite element simulations, using the material models derived in Section \ref{sec:material_characterization}). These labels represent the quantities of interest in the inference process (e.g., the contact force distribution). The following subsections describe in detail each of the components of the dataset.

\subsection{Features} \label{sec:features}
\begin{figure}
	\centering
	\includegraphics[width=0.9\linewidth]{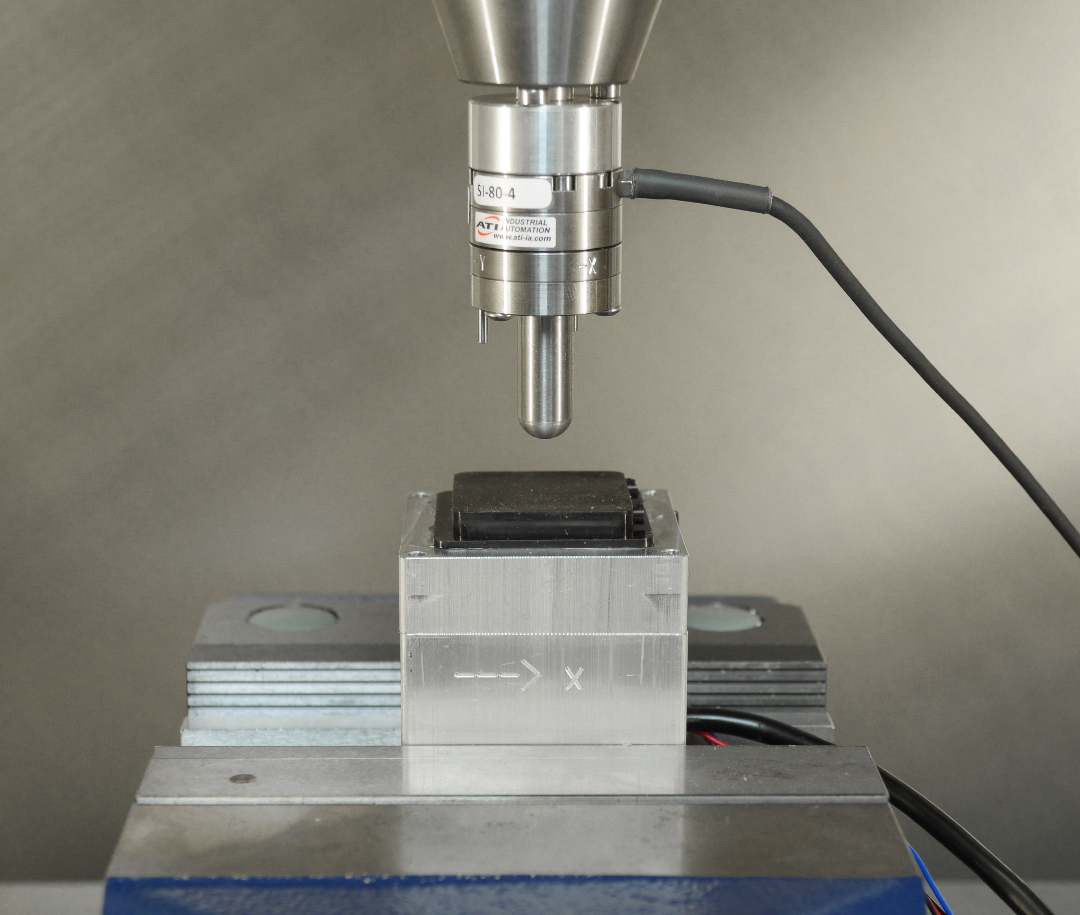}
	\caption{The experimental data collection setup is shown above. The indenter and the F/T sensor (connected through the cable on the top right) are attached to the spindle of an automatic milling machine.}
	\label{fig:milling_machine}
\end{figure}

In order to perform a large number of indentations within a feasible time, an automatic milling and drilling machine (Fehlmann PICOMAX 56 TOP) is used to press an indenter against the soft surface of the tactile sensor at different locations and depths. The machine is equipped with fast and precise motion control (up to $10^{-3}$ mm). In the experiments presented here, a stainless steel spherical-ended cylindrical indenter is used. The indenter has a diameter of 10 mm and is attached to the spindle of the milling machine, together with a six-axis F/T sensor (ATI Mini 27 Titanium). The experimental data collection setup is shown in Fig. \ref{fig:milling_machine}.

\begin{figure}
	\centering
	\subfloat[Sample indentation]{%
		\includegraphics[width=1\linewidth]{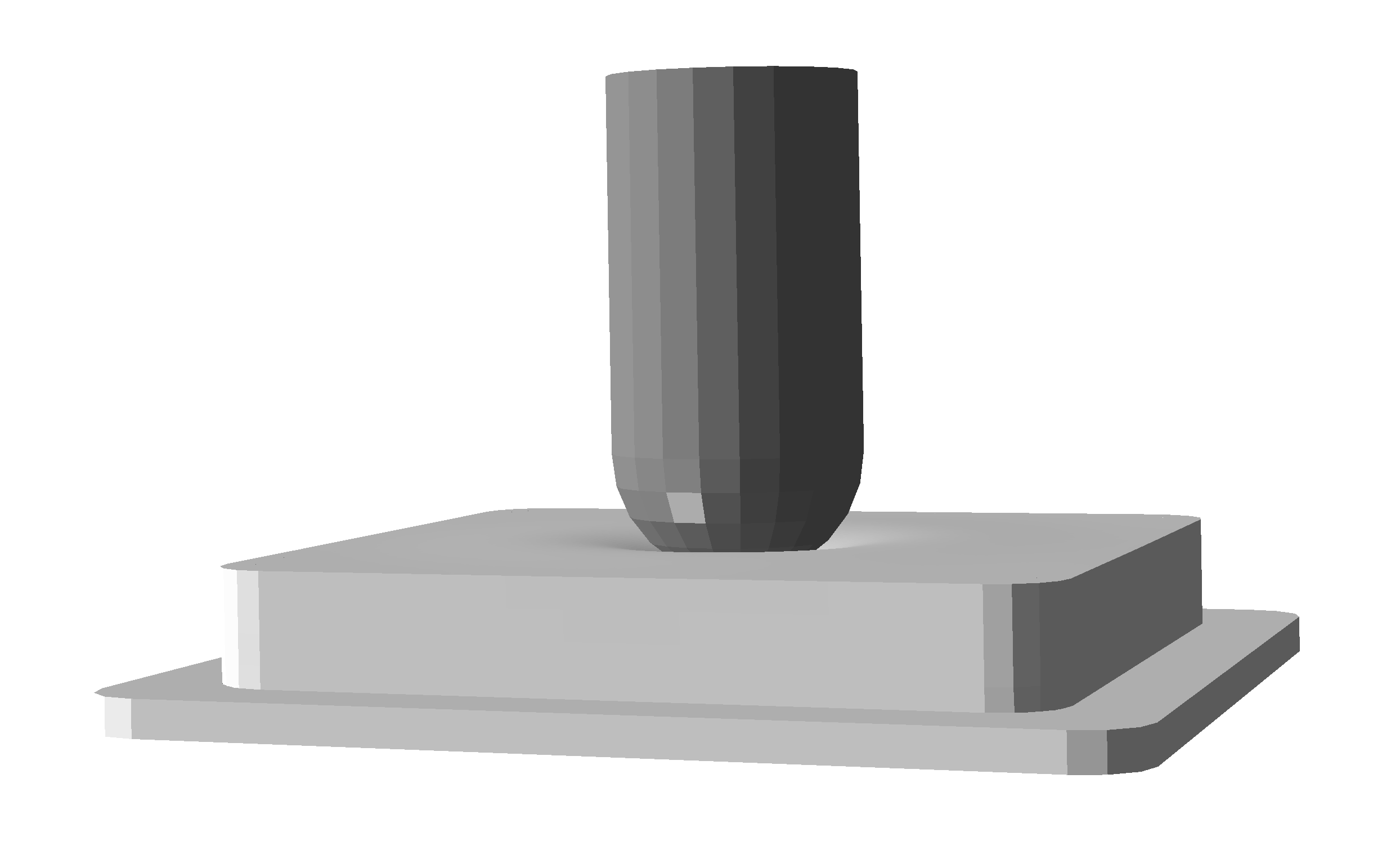} }
	\hfill
	\subfloat[Contact pressure distribution]{%
		\includegraphics[width=0.6\linewidth]{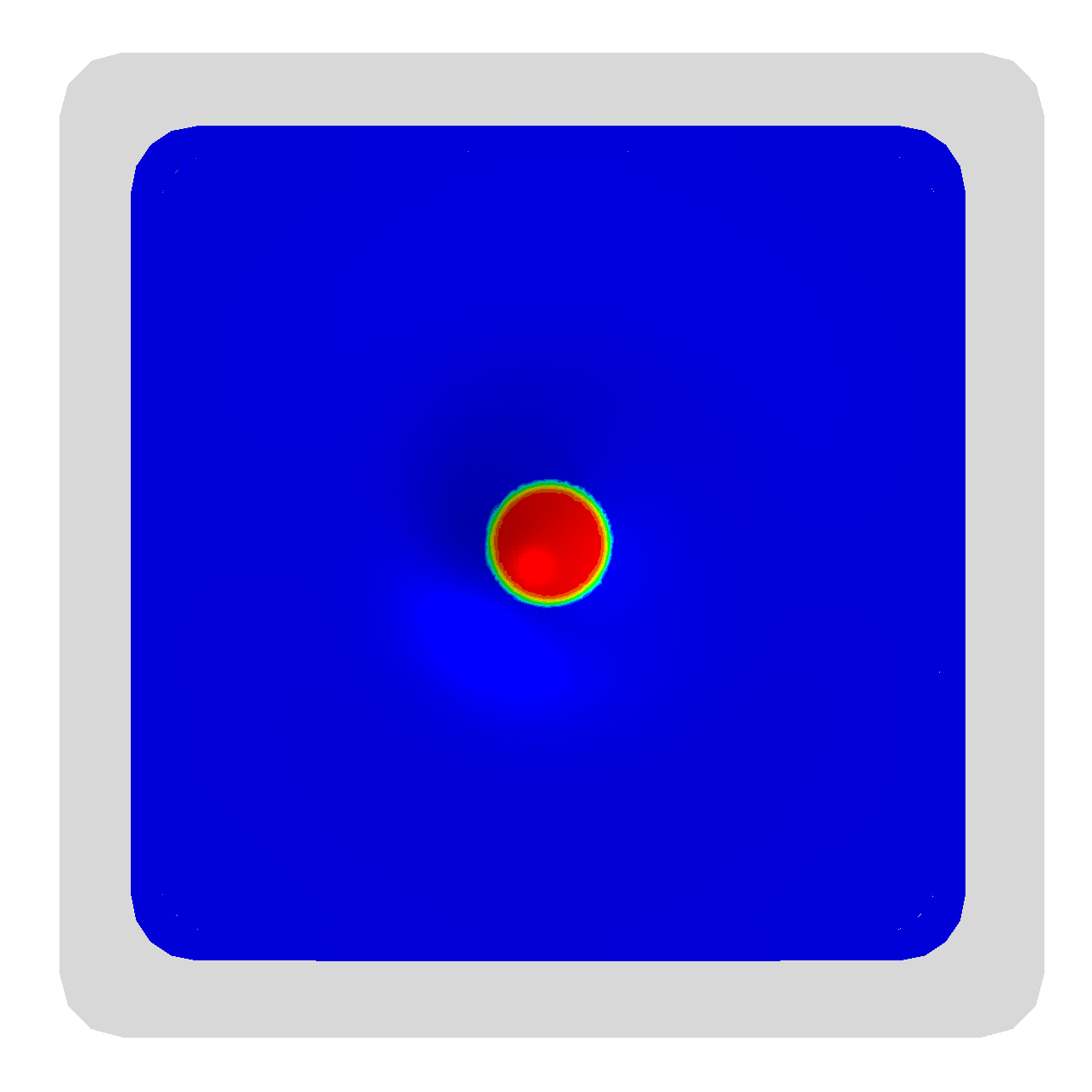} }
	\caption{The result of a sample FEM indentation in Abaqus is shown in this figure. The indenter and the gel are modeled to reflect their actual material and geometric properties, see (a). An example of the resulting contact pressure distribution (top view) is shown in (b), where the colors are mapped to the pressure magnitude (from zero, in blue, to the maximum, in red). }
	\label{fig:fem_sim}
\end{figure}

A total of 13,448 vertical indentations were performed, on an horizontal grid with a regular spacing of 0.55 mm, at various depths (with a maximum depth of 2 mm). The RGB images of the particle spread are captured once the indentation has reached the commanded position. Five images were collected for each indentation in order to improve the robustness to image noise. The F/T sensor's measurements of the total vertical and horizontal (over two perpendicular axes) contact force were recorded. 

The optical flow field is extracted from the images (converted to grayscale) through an algorithm based on Dense Inverse Search \cite{dis_paper}. The magnitude and the direction of the field are then averaged in $m$ image regions of equal area, as described in \cite{sferrazza_sensors}. The tuples of magnitude and direction for each of these regions yield a set of $2 \times m$ features for each data point.

The readings from the F/T sensor are used to assess the quality of the ground truth labels, as described in the next subsection. The range of forces recorded in this procedure spans up to 1.7 N in the vertical direction and 0.15 N in each of the horizontal axes. Note that the large difference in magnitude between the vertical and horizontal forces is mainly due to the symmetry of the indentations, which leads to the cancellation of the symmetric contributions to the total horizontal force, with the exception of the regions close to the edges of the surface.

\subsection{Labels}
Although the F/T sensor provides the total contact force in each direction, it does not provide any information about the force distribution over the contact surface. The force distribution renders a compact representation of various contact aspects for generic indentations. In fact, it encodes information about the contact area and forces applied to the surface, even in the case of interactions with objects of complex geometries or when multiple and distinct contact points are present. An example application, in which both the contact area and the total contact force are necessary, is presented in \cite{bolt_insertion}, where the interaction with a specific object is exploited. The contact force distribution is obtained in this article through FEA, which essentially simulates the indentation experiments performed with the milling machine, see for example Fig. \ref{fig:fem_sim}.

The FEM simulations are carried out in Abaqus/Standard \cite{abaqus_manual}. The geometry of the two top layers of the sensor is modeled as shown in Fig. \ref{fig:three_layers}, and material properties are assigned to each layer as described in Section \ref{sec:material_characterization}, using the implementation of Ogden's model provided by Abaqus. Note that this neglects the influence of the spherical particles, as the model was derived from tests on the pure Ecoflex GEL. Assuming rigid particles, the ratio between the Ecoflex--particle composite modulus $E_\mathrm{c}$ and the Young's modulus $E$ of the pure Ecoflex GEL can be estimated using Eshelby inclusion theory \cite{eshelby_inclusions}, $E_\mathrm{c}/E = 1/(1-5\phi/2) \approx 1.05$ for a particle volume fraction $\phi = 0.0196$. The spherical-ended indenter is modeled as an analytical rigid shell. The finite element mesh is composed of linear tetrahedral elements (C3D4H) and hexahedral elements with reduced integration (C3D8RH). Both element types are used with hybrid formulation as appropriate for incompressible materials. A local mesh refinement is performed at the contact and at the interface of the materials, with a characteristic element size of 0.3~mm. Tie constraints are applied at the material interface to enforce the same displacement of the nodes in contact. The bottom nodes are fixed, reflecting the interface with the much stiffer bottom layer (Elastosil 7:1).

The contact between the top surface and the indenter is modeled as a hard contact and discretized with a surface-to-surface method. The friction coefficient between the indenter and the top layer is estimated by letting a block of Elastosil 25:1 rest on an inclined stainless steel plate. The maximum tilt angle $\theta$ before the block begins to slide is recorded with an external camera, and the static friction coefficient $\mu_0$ is calculated from static equilibrium as $\mu_0 = \tan{\theta}$. This procedure yields a friction coefficient of 0.45. The friction coefficient was assumed constant, a possible dependence on the contact pressure was neglected.

The FEM simulations generate the normal and shear components of the contact force distribution resulting from each indentation. Note that both the normal and shear forces acting at each node are generally 3D vectors. As an example, the normal force that stems from a pure vertical indentation is not necessarily vertical, as a consequence of the material deformation (although the vertical component generally has the largest magnitude).

\begin{figure}
	\scalebox{0.45}{\tikzstyle{ipe stylesheet} = [
  ipe import,
  even odd rule,
  line join=round,
  line cap=butt,
  ipe pen normal/.style={line width=0.4},
  ipe pen heavier/.style={line width=0.8},
  ipe pen fat/.style={line width=1.2},
  ipe pen ultrafat/.style={line width=2},
  ipe pen normal,
  ipe mark normal/.style={ipe mark scale=3},
  ipe mark large/.style={ipe mark scale=5},
  ipe mark small/.style={ipe mark scale=2},
  ipe mark tiny/.style={ipe mark scale=1.1},
  ipe mark normal,
  /pgf/arrow keys/.cd,
  ipe arrow normal/.style={scale=7},
  ipe arrow large/.style={scale=10},
  ipe arrow small/.style={scale=5},
  ipe arrow tiny/.style={scale=3},
  ipe arrow normal,
  /tikz/.cd,
  ipe arrows, 
  <->/.tip = ipe normal,
  ipe dash normal/.style={dash pattern=},
  ipe dash dashed/.style={dash pattern=on 4bp off 4bp},
  ipe dash dotted/.style={dash pattern=on 1bp off 3bp},
  ipe dash dash dotted/.style={dash pattern=on 4bp off 2bp on 1bp off 2bp},
  ipe dash dash dot dotted/.style={dash pattern=on 4bp off 2bp on 1bp off 2bp on 1bp off 2bp},
  ipe dash normal,
  ipe node/.append style={font=\normalsize},
  ipe stretch normal/.style={ipe node stretch=1},
  ipe stretch normal,
  ipe opacity 10/.style={opacity=0.1},
  ipe opacity 30/.style={opacity=0.3},
  ipe opacity 50/.style={opacity=0.5},
  ipe opacity 75/.style={opacity=0.75},
  ipe opacity opaque/.style={opacity=1},
  ipe opacity opaque,
]
\definecolor{red}{rgb}{1,0,0}
\definecolor{green}{rgb}{0,1,0}
\definecolor{blue}{rgb}{0,0,1}
\definecolor{yellow}{rgb}{1,1,0}
\definecolor{orange}{rgb}{1,0.647,0}
\definecolor{gold}{rgb}{1,0.843,0}
\definecolor{purple}{rgb}{0.627,0.125,0.941}
\definecolor{gray}{rgb}{0.745,0.745,0.745}
\definecolor{brown}{rgb}{0.647,0.165,0.165}
\definecolor{navy}{rgb}{0,0,0.502}
\definecolor{pink}{rgb}{1,0.753,0.796}
\definecolor{seagreen}{rgb}{0.18,0.545,0.341}
\definecolor{turquoise}{rgb}{0.251,0.878,0.816}
\definecolor{violet}{rgb}{0.933,0.51,0.933}
\definecolor{darkblue}{rgb}{0,0,0.545}
\definecolor{darkcyan}{rgb}{0,0.545,0.545}
\definecolor{darkgray}{rgb}{0.663,0.663,0.663}
\definecolor{darkgreen}{rgb}{0,0.392,0}
\definecolor{darkmagenta}{rgb}{0.545,0,0.545}
\definecolor{darkorange}{rgb}{1,0.549,0}
\definecolor{darkred}{rgb}{0.545,0,0}
\definecolor{lightblue}{rgb}{0.678,0.847,0.902}
\definecolor{lightcyan}{rgb}{0.878,1,1}
\definecolor{lightgray}{rgb}{0.827,0.827,0.827}
\definecolor{lightgreen}{rgb}{0.565,0.933,0.565}
\definecolor{lightyellow}{rgb}{1,1,0.878}
\definecolor{black}{rgb}{0,0,0}
\definecolor{white}{rgb}{1,1,1}
\begin{tikzpicture}[ipe stylesheet]
  \draw[red, ipe pen ultrafat]
    (320, 768) rectangle (576, 512);
  \draw
    (352, 736)
     -- (432, 688)
     -- (432, 736)
     -- (352, 736)
     -- (352, 608)
     -- (432, 688)
     -- (464, 656)
     -- (400, 656)
     -- (448, 608)
     -- (352, 608)
     -- (416, 544)
     -- (448, 608)
     -- (512, 576)
     -- (416, 544)
     -- (512, 480)
     -- (464, 560)
     -- (544, 528)
     -- (512, 576)
     -- (528, 624)
     -- (448, 608)
     -- (464, 656)
     -- (528, 624)
     -- (544, 672)
     -- (464, 656)
     -- (512, 704)
     -- (432, 688)
     -- (432, 688);
  \draw
    (432, 736)
     -- (512, 704)
     -- (544, 672)
     -- (608, 608)
     -- (528, 624)
     -- (544, 528)
     -- (608, 608)
     -- (608, 608);
  \draw
    (544, 528)
     -- (592, 480)
     -- (608, 608)
     -- (560, 704)
     -- (544, 672)
     -- (544, 672);
  \draw
    (512, 704)
     -- (560, 704)
     -- (608, 720)
     -- (608, 608);
  \draw
    (512, 704)
     -- (512, 752)
     -- (432, 736)
     -- (384, 784)
     -- (352, 736)
     -- (288, 672)
     -- (352, 608)
     -- (336, 528)
     -- (416, 544)
     -- (368, 496)
     -- (336, 528)
     -- (288, 672)
     -- (272, 768)
     -- (352, 736)
     -- (336, 800)
     -- (272, 768)
     -- (272, 768);
  \draw
    (336, 800)
     -- (384, 784)
     -- (512, 752)
     -- (496, 800)
     -- (336, 800)
     -- (496, 800);
  \draw
    (512, 752)
     -- (560, 704)
     -- (576, 800)
     -- (512, 752);
  \draw
    (608, 720)
     -- (576, 800);
  \draw
    (496, 800)
     -- (576, 800)
     -- (576, 800);
  \draw
    (368, 496)
     -- (288, 496)
     -- (336, 528);
  \draw
    (288, 496)
     -- (288, 544)
     -- (336, 528);
  \draw
    (288, 544)
     -- (288, 672);
  \draw
    (368, 496)
     -- (512, 480);
  \draw
    (512, 480)
     -- (544, 528);
  \draw
    (512, 480)
     -- (592, 480);
  \draw
    (432, 688)
     -- (352, 672);
  \draw
    (288, 672)
     -- (352, 672);
  \draw
    (464, 560)
     -- (448, 608)
     -- (448, 608);
  \draw
    (400, 656)
     -- (352, 672);
  \draw[shift={(576, 768)}, yscale=0.5, red, ipe pen ultrafat, ipe dash dashed]
    (0, 0)
     -- (0, 96);
  \draw[shift={(576, 768)}, xscale=0.5, red, ipe pen ultrafat, ipe dash dashed]
    (0, 0)
     -- (96, 0);
  \draw[shift={(576, 512)}, yscale=0.5, red, ipe pen ultrafat, ipe dash dashed]
    (0, 0)
     -- (0, -96);
  \draw[shift={(576, 512)}, xscale=0.5, red, ipe pen ultrafat, ipe dash dashed]
    (0, 0)
     -- (96, 0);
  \draw[shift={(320, 512)}, xscale=0.5, red, ipe pen ultrafat, ipe dash dashed]
    (0, 0)
     -- (-96, 0);
  \draw[shift={(320, 512)}, yscale=0.5, red, ipe pen ultrafat, ipe dash dashed]
    (0, 0)
     -- (0, -96);
  \draw[shift={(320, 768)}, yscale=0.5, red, ipe pen ultrafat, ipe dash dashed]
    (0, 0)
     -- (0, 96);
  \draw[shift={(320, 768)}, xscale=0.5, red, ipe pen ultrafat, ipe dash dashed]
    (0, 0)
     -- (-96, 0);
  \draw
    (384, 784)
     -- (496, 800);
  \draw[ipe dash dashed]
    (288, 672)
     -- (272, 656);
  \draw[ipe dash dashed]
    (288, 544)
     -- (272, 560);
  \draw[ipe dash dashed]
    (288, 496)
     -- (272, 464);
  \draw[ipe dash dashed]
    (272, 768)
     -- (288, 816);
  \draw[ipe dash dashed]
    (336, 800)
     -- (320, 816);
  \draw[ipe dash dashed]
    (496, 800)
     -- (512, 816);
  \draw[ipe dash dashed]
    (576, 800)
     -- (624, 752);
  \draw[ipe dash dashed]
    (608, 720)
     -- (624, 736);
  \draw[ipe dash dashed]
    (608, 608)
     -- (624, 560);
  \draw[ipe dash dashed]
    (592, 480)
     -- (608, 496);
  \draw[ipe dash dashed]
    (608, 496)
     -- (624, 512);
  \draw[ipe dash dashed]
    (592, 480)
     -- (560, 464);
  \draw[ipe dash dashed]
    (512, 480)
     -- (528, 464);
  \draw[ipe dash dashed]
    (512, 480)
     -- (464, 464);
  \draw[ipe dash dashed]
    (368, 496)
     -- (400, 464);
  \draw[ipe dash dashed]
    (368, 496)
     -- (304, 464);
  \pic[ipe mark large, green]
     at (352, 736) {ipe disk};
  \pic[ipe mark large, green]
     at (352, 672) {ipe disk};
  \pic[ipe mark large, green]
     at (352, 608) {ipe disk};
  \pic[ipe mark large, green]
     at (336, 528) {ipe disk};
  \pic[ipe mark large, green]
     at (416, 544) {ipe disk};
  \pic[ipe mark large, green]
     at (448, 608) {ipe disk};
  \pic[ipe mark large, green]
     at (400, 656) {ipe disk};
  \pic[ipe mark large, green]
     at (432, 688) {ipe disk};
  \pic[ipe mark large, green]
     at (432, 736) {ipe disk};
  \pic[ipe mark large, green]
     at (512, 752) {ipe disk};
  \pic[ipe mark large, green]
     at (512, 704) {ipe disk};
  \pic[ipe mark large, green]
     at (464, 656) {ipe disk};
  \pic[ipe mark large, green]
     at (528, 624) {ipe disk};
  \pic[ipe mark large, green]
     at (544, 672) {ipe disk};
  \pic[ipe mark large, green]
     at (560, 704) {ipe disk};
  \pic[ipe mark large, green]
     at (512, 576) {ipe disk};
  \pic[ipe mark large, green]
     at (464, 560) {ipe disk};
  \pic[ipe mark large, green]
     at (544, 528) {ipe disk};
  \filldraw[ipe dash dashed, fill=green]
    (592, 480)
     -- (608, 464);
\end{tikzpicture}}
	\caption{The procedure to discretize the force distribution is sketched in this figure. The sensor's surface is discretized into $n$ bins of equal size, with the boundaries shown in red. The surface mesh used for the FEA is shown in black in the undeformed state. Each node is assigned to the bin that contains it, as is the case for the green nodes contained in the bin shown in solid red. For each indentation, the resulting forces at the nodes assigned to the same bin are summed along each axis to determine the three label vector components for the corresponding bin.}
	\label{fig:discretization}
\end{figure}
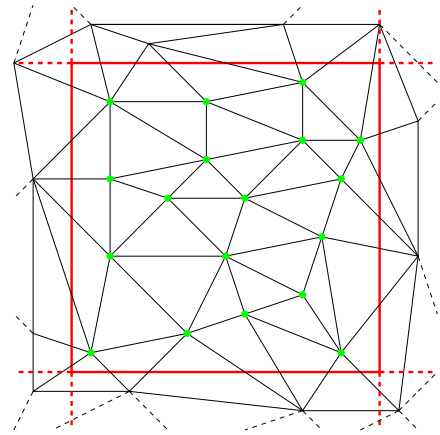

The normal and shear force distributions are discretized by summing the respective nodal forces inside $n$ surface bins, as shown in Fig. \ref{fig:discretization}. The resulting 3D force for each of these bins is used as a ground truth label with $3 \times n$ components for each data point.

This procedure is applied to assign ground truth labels to the 13,448 indentations described in the previous subsection. Since there are no readily available sensors that measure the full contact force distribution with high spatial resolution and without altering the sensor's soft surface, the quality of the labels is evaluated by comparing the components of the total force resulting from the FEM simulations with the ones measured by the F/T sensor. Note that the total force components can be obtained from the FEM simulations by summing the contact force values of all the $n$ bins, or simply summing the force values at all nodes of the surface mesh used for the FEA. The resulting root-mean-square error on the ground truth ($\text{RMSE}_\text{GT}$) for the entire dataset is reported in Table \ref{table:agreement_fem_ft} for each component. $x$ and $y$ are the horizontal axes, and $z$ is the vertical axis, which is positive pointing from the camera towards the top surface. The resulting errors are comparable to the F/T sensor's resolution, shown in the table as a reference. In Fig. \ref{fig:agreement}, the plots show the agreement on the $z$ component of the total force between the F/T sensor's readings and the results from the FEA for two of the indentation locations. The good agreement between the F/T measurements and the FEM simulations further justifies the simplifying assumptions taken in the material characterization and the FEM modeling. Additionally, the same plots show that using a linear elastic material model and neglecting geometric nonlinearities (i.e., NLgeom flag in Abaqus) lead to a considerable performance loss for large deformations.

\begin{table}[h!]
	\centering
	\caption{Total force agreement (FEA vs F/T sensor)}
	\label{table:agreement_fem_ft}
	\begin{tabular}{c|c|c} 
		\hline
		\rule{0pt}{2ex}
		Axis & $\text{RMSE}_\text{GT}$ & F/T resolution\\
		\hline
		\rule{0pt}{3ex}
		$x$ & 0.02 N & 0.03 N\\ 
		$y$ & 0.02 N & 0.03 N\\
		$z$ & 0.06 N & 0.06 N\\
		\hline
	\end{tabular}
\end{table}

\newlength{\figureheight}
\newlength{\figurewidth}

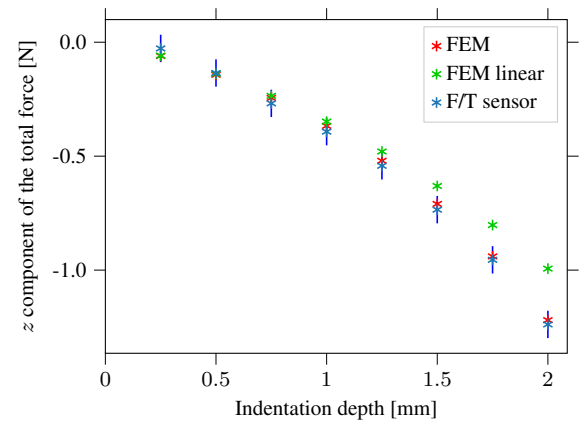
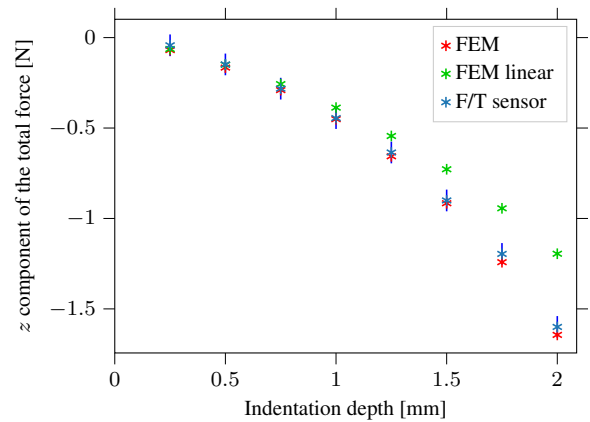
\begin{figure}
	\centering
	\setlength\figureheight{0.7\columnwidth}
	\setlength\figurewidth{0.9\columnwidth}
	\subfloat[Center indentation]{%
\begin{tikzpicture}

\definecolor{color0}{rgb}{0.12156862745098,0.466666666666667,0.705882352941177}

\begin{axis}[
height=\figureheight,
legend cell align={left},
legend style={draw=white!80.0!black},
tick align=outside,
tick pos=both,
width=\figurewidth,
x grid style={white!69.01960784313725!black},
xlabel={Indentation depth [mm]},
xmin=0, xmax=2.0875,
xtick style={color=black},
y grid style={white!69.01960784313725!black},
ylabel={$z$ component of the total force [N]},
ymin=-1.3649556210528, ymax=0.0992955934674039,
ytick style={color=black},
ytick={-1.0,-0.5,0.0},
yticklabels={-1.0,-0.5,0.0}
]
\path [draw=blue, semithick]
(axis cs:0.25,-0.0872612799198783)
--(axis cs:0.25,0.0327387200801217);

\path [draw=blue, semithick]
(axis cs:0.5,-0.195030834304619)
--(axis cs:0.5,-0.0750308343046191);

\path [draw=blue, semithick]
(axis cs:0.75,-0.328243535736672)
--(axis cs:0.75,-0.208243535736672);

\path [draw=blue, semithick]
(axis cs:1,-0.452540224779301)
--(axis cs:1,-0.332540224779301);

\path [draw=blue, semithick]
(axis cs:1.25,-0.602338756296482)
--(axis cs:1.25,-0.482338756296482);

\path [draw=blue, semithick]
(axis cs:1.5,-0.794801171067046)
--(axis cs:1.5,-0.674801171067046);

\path [draw=blue, semithick]
(axis cs:1.75,-1.01485703606129)
--(axis cs:1.75,-0.894857036061292);

\path [draw=blue, semithick]
(axis cs:2,-1.29839874766552)
--(axis cs:2,-1.17839874766552);

\addplot [semithick, red, mark=asterisk, mark size=2, mark options={solid}, only marks]
table {%
0.25 -0.060427
0.5 -0.14277
0.75 -0.24445
1 -0.36741
1.25 -0.52029
1.5 -0.70962
1.75 -0.93993
2 -1.2196
};
\addlegendentry{FEM}
\addplot [semithick, green!80.0!black, mark=asterisk, mark size=2, mark options={solid}, only marks]
table {%
0.25 -0.058655
0.5 -0.13866
0.75 -0.23547
1 -0.34826
1.25 -0.4796
1.5 -0.6308
1.75 -0.80231
2 -0.99344
};
\addlegendentry{FEM linear}
\addplot [semithick, color0, mark=asterisk, mark size=2, mark options={solid}, only marks]
table {%
0.25 -0.0272612799198783
0.5 -0.135030834304619
0.75 -0.268243535736672
1 -0.392540224779301
1.25 -0.542338756296482
1.5 -0.734801171067046
1.75 -0.954857036061292
2 -1.23839874766552
};
\addlegendentry{F/T sensor}
\end{axis}

\end{tikzpicture} }
	\hfill
	\subfloat[Corner indentation]{%
\begin{tikzpicture}

\definecolor{color0}{rgb}{0.12156862745098,0.466666666666667,0.705882352941177}

\begin{axis}[
height=\figureheight,
legend cell align={left},
legend style={draw=white!80.0!black},
tick align=outside,
tick pos=both,
width=\figurewidth,
x grid style={white!69.01960784313725!black},
xlabel={Indentation depth [mm]},
xmin=0, xmax=2.0875,
xtick style={color=black},
y grid style={white!69.01960784313725!black},
ylabel={$z$ component of the total force [N]},
ymin=-1.74364291541733, ymax=0.101332698813621,
ytick style={color=black}
]
\path [draw=blue, semithick]
(axis cs:0.25,-0.102529829105968)
--(axis cs:0.25,0.017470170894032);

\path [draw=blue, semithick]
(axis cs:0.5,-0.208304668545078)
--(axis cs:0.5,-0.0883046685450775);

\path [draw=blue, semithick]
(axis cs:0.75,-0.342261288973283)
--(axis cs:0.75,-0.222261288973283);

\path [draw=blue, semithick]
(axis cs:1,-0.505039052934356)
--(axis cs:1,-0.385039052934356);

\path [draw=blue, semithick]
(axis cs:1.25,-0.695234601799479)
--(axis cs:1.25,-0.575234601799479);

\path [draw=blue, semithick]
(axis cs:1.5,-0.960406083673237)
--(axis cs:1.5,-0.840406083673237);

\path [draw=blue, semithick]
(axis cs:1.75,-1.25606442250167)
--(axis cs:1.75,-1.13606442250167);

\path [draw=blue, semithick]
(axis cs:2,-1.65978038749774)
--(axis cs:2,-1.53978038749774);

\addplot [semithick, red, mark=asterisk, mark size=2, mark options={solid}, only marks]
table {%
0.25 -0.068949
0.5 -0.16576
0.75 -0.29019
1 -0.45087
1.25 -0.65615
1.5 -0.91577
1.75 -1.2417
2 -1.6438
};
\addlegendentry{FEM}
\addplot [semithick, green!80.0!black, mark=asterisk, mark size=2, mark options={solid}, only marks]
table {%
0.25 -0.0631
0.5 -0.15015
0.75 -0.25716
1 -0.38721
1.25 -0.54398
1.5 -0.72787
1.75 -0.94382
2 -1.1951
};
\addlegendentry{FEM linear}
\addplot [semithick, color0, mark=asterisk, mark size=2, mark options={solid}, only marks]
table {%
0.25 -0.042529829105968
0.5 -0.148304668545078
0.75 -0.282261288973283
1 -0.445039052934356
1.25 -0.635234601799479
1.5 -0.900406083673237
1.75 -1.19606442250167
2 -1.59978038749774
};
\addlegendentry{F/T sensor}
\end{axis}

\end{tikzpicture} }
	\caption{The plots above show the agreement on the total vertical contact force between the measurements obtained from the F/T sensor (in blue) and the FEM simulations (in red). The results from the simulations are accurate for indentations at the center of the surface (a) and close to the corners (b) (5 mm from each of the edges) for different indentation depths. The F/T sensor readings are shown with $\pm 0.06$ N bars, representing the resolution of the F/T sensor. In green, the results obtained using a linear elastic model (as opposed to the hyperelastic model described in Section \ref{sec:material_characterization}) and neglecting geometric nonlinearities are shown.}
	\label{fig:agreement}
\end{figure}

Although the FEM simulations can be time consuming to carry out, depending on the accuracy required, most of the operations are highly parallelizable, as for example, the several indentations. This makes it possible to exploit cluster computers or GPUs to reduce the time consumption. The simulations presented here are carried out on the Euler cluster of ETH Zurich.

Note that the strategy presented above provides the ground truth for the full contact force distribution under no assumptions on the specific tactile sensing technique.  It is therefore not limited to use on vision-based devices, but more generally on data-driven approaches to the force reconstruction task.

\section{Neural network training} \label{sec:learning}
\subsection{Learning architecture}
\begin{figure}
	\scalebox{0.25}{\input{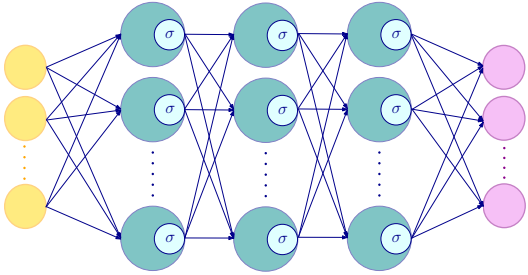}}
	\caption{A diagram of the learning architecture used to predict the 3D contact force distribution. In yellow the input layer, representing the image features, in cyan the hidden layers, and in magenta the output layer, representing the discretized force distribution.}
	\label{fig:learning_architecture}
\end{figure}

A feedforward DNN architecture (see Fig. \ref{fig:learning_architecture}) is used to address the supervised learning task of reconstructing the full contact force distribution from the features extracted from the images. An input layer with $2\times m$ units represents the image features described in Section \ref{sec:features} (a tuple of averaged optical flow magnitude and direction for each of the chosen image regions). Similarly, an output layer with $3 \times n$ units represents the discretized force distribution applied to the surface of the sensor (a three-dimensional force vector for each of the discrete surface bins).

Three fully connected hidden layers with a sigmoid activation function are used to model the map between the inputs and the outputs. Dropout layers are used after each of the hidden layers during the training phase. Twenty percent of the dataset is used as a test set, while the remaining data are used for training. The architecture is trained with PyTorch\footnote{\url{www.pytorch.org}} by minimizing the mean squared error (MSE) through the Adam optimizer, see \cite{adam_optimizer}. The remaining parameters chosen for the optimization, as well as the size of each layer, are summarized in Table \ref{table:parameters_dnn}. Note that the spatial resolution of the tactile sensor is determined by the size of the surface bins, which have a side of 1.6 mm, comparable to the spatial resolution of the human fingertip \cite{human_fingertip}. However, a finer resolution may yield additional discrimination capabilities, i.e., for sensing an object's roughness or texture.

In contrast to architectures that directly learn from the pixels (e.g., convolutional neural networks), the extraction of the optical flow features described in Section \ref{sec:features} yields a relatively shallow model, reducing the training times and the data requirements. Additionally, the use of these features makes it possible to efficiently transfer the model across different sensors, as shown in previous work \cite{transfer_learning_sferrazza}. However, these features do not exploit the information about particles placed at different distances from the camera.

\begin{table}[h!]
	\centering
	\caption{DNN parameters}
	\label{table:parameters_dnn}
	\begin{tabular}{c|c|c} 
		\hline
		\rule{0pt}{2ex}
		Symbol & Value & Description\\
		\hline
		\rule{0pt}{3ex}
		$m$ & 1600 & \# of averaging image regions\\ 
		$n$ & 400 & \# of discrete surface bins\\
		- & (800, 600, 400) & hidden layers' size \\
		- & 1E-4 & learning rate \\
		- & 400 & training batch size \\
		- & 0.1 & dropout rate \\
		\hline
	\end{tabular}
\end{table}

\subsection{Results}
After training, the quality of the DNN predictions is evaluated on the test set. Additionally to the root-mean-square error (RMSE) on the entire test set, the sparse RMSE on the non-zero values of the FEM ground truth is also computed as

\begin{align*}
\text{RMSES} := \sqrt{\frac{1}{|\mathcal{I}|} \sum_{(i,l)\in \mathcal{I}} \left(f_i^{(l)} - \hat{f}_i^{(l)}\right)^2 },
\end{align*}
where $f_i^{(l)}$ and $\hat{f}_i^{(l)}$ are the $i$-th components of the ground truth and the predicted label, respectively, for the $l$-th sample in the test set, and,
\begin{align*}
\mathcal{I} \! := \! \left\{ (i,l)\! \in \!\{0,\dots,3n-1\}\! \times \! \{0,\dots,N_\text{set}-1\} \mid f_i^{(l)} \neq 0 \right\},
\end{align*}
with $N_\text{set}$ the number of samples in the test set. This metric emphasizes the prediction performance in the location where the contact is expected.

\begin{figure*}
	\centering
	\subfloat[$x$ component of the predicted force distribution]{%
	\includegraphics[width=0.5\textwidth]{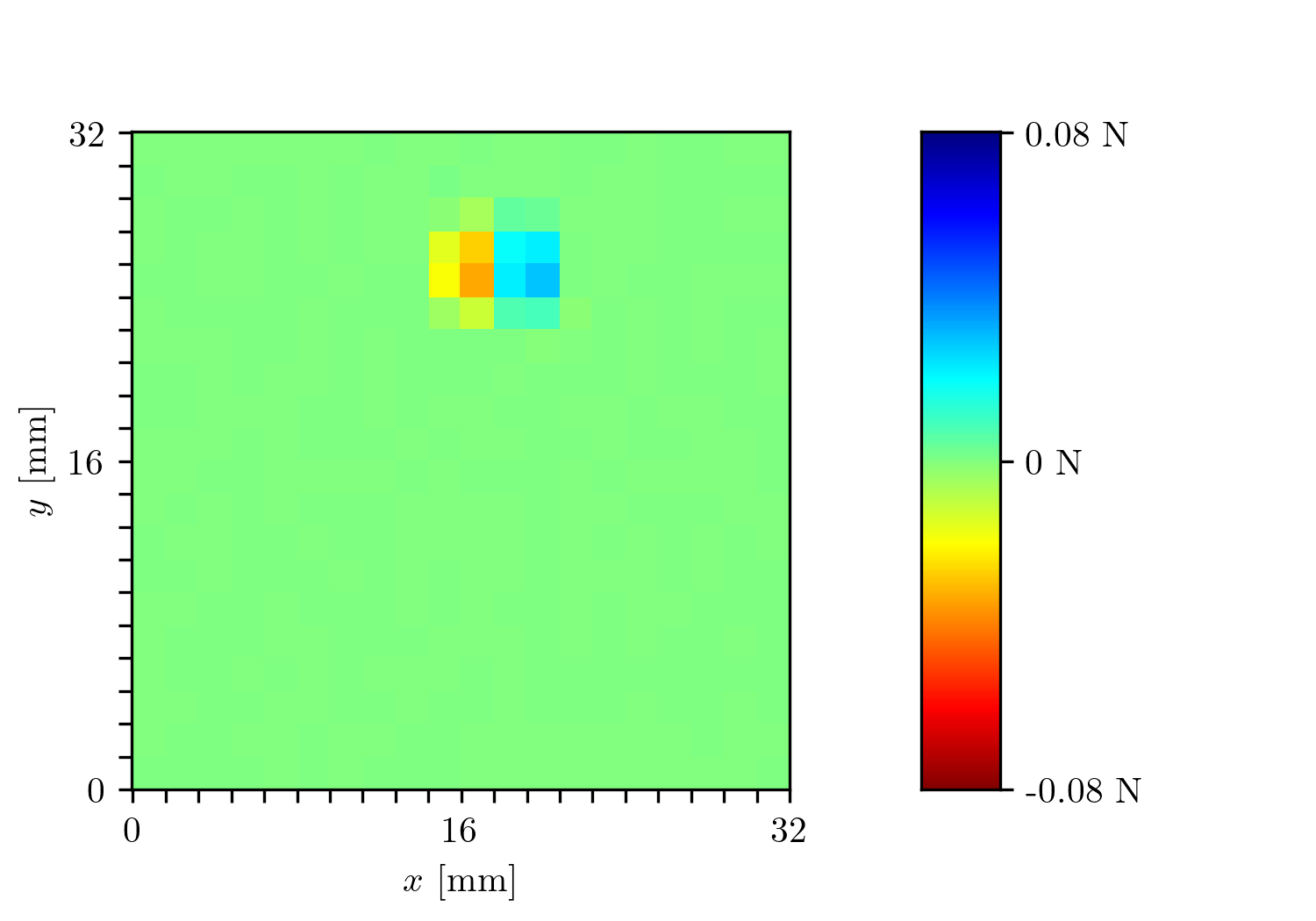} }
	\subfloat[$x$ component of the ground truth force distribution]{%
	\includegraphics[width=0.5\linewidth]{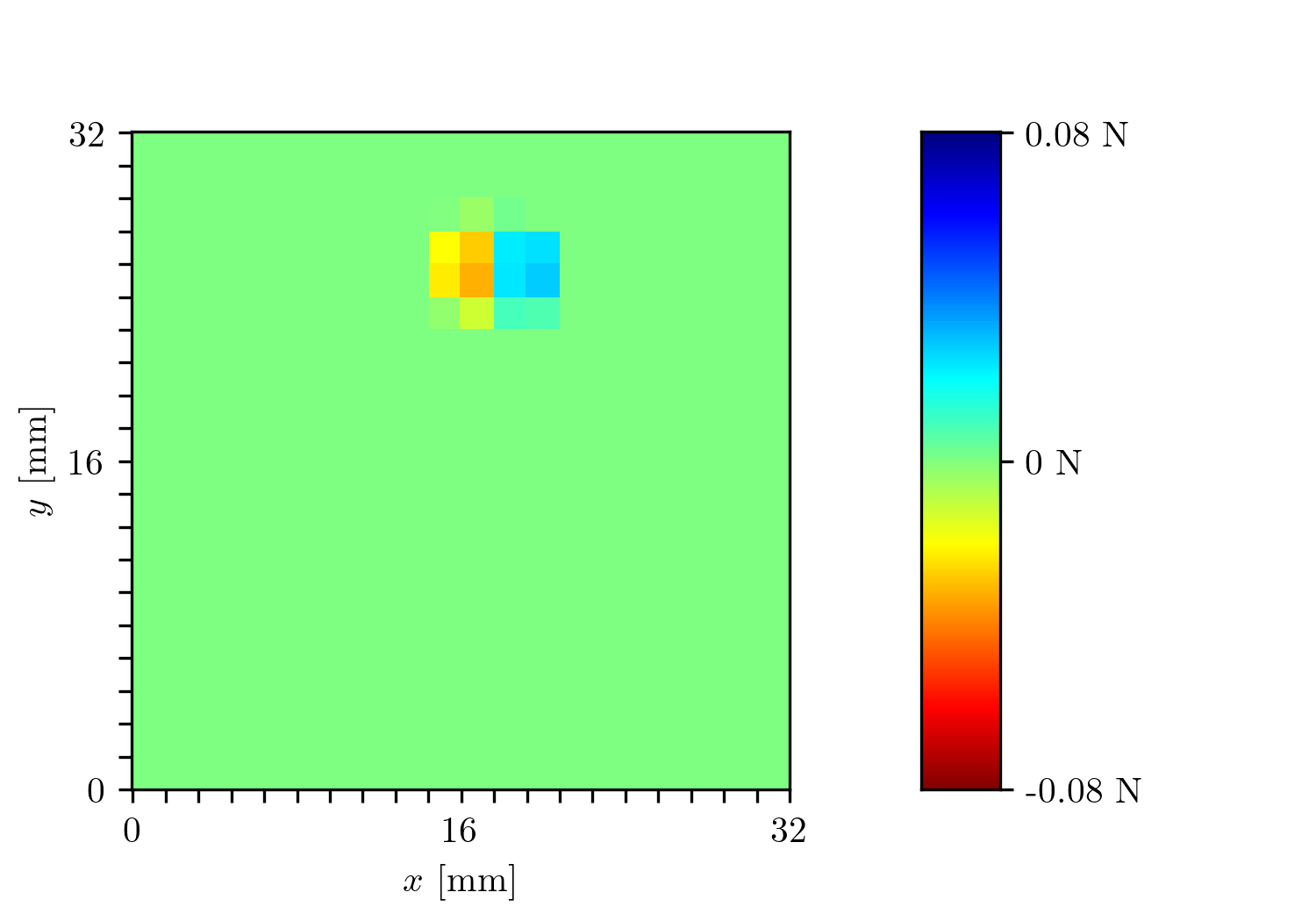} }
	\hfill
	\centering
	\subfloat[$y$ component of the predicted force distribution]{%
	\includegraphics[width=0.5\textwidth]{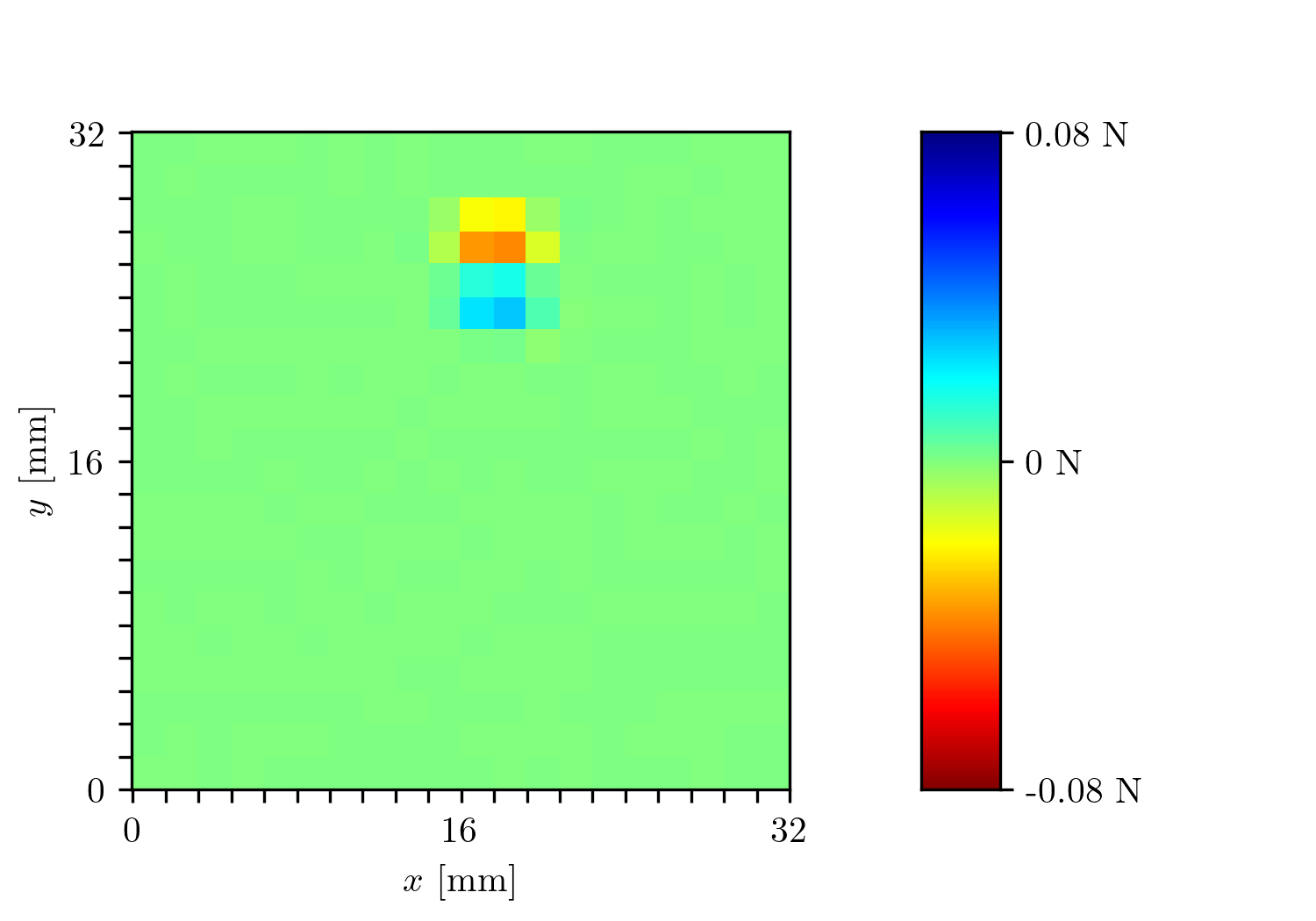} }
	\subfloat[$y$ component of the ground truth force distribution]{%
	\includegraphics[width=0.5\linewidth]{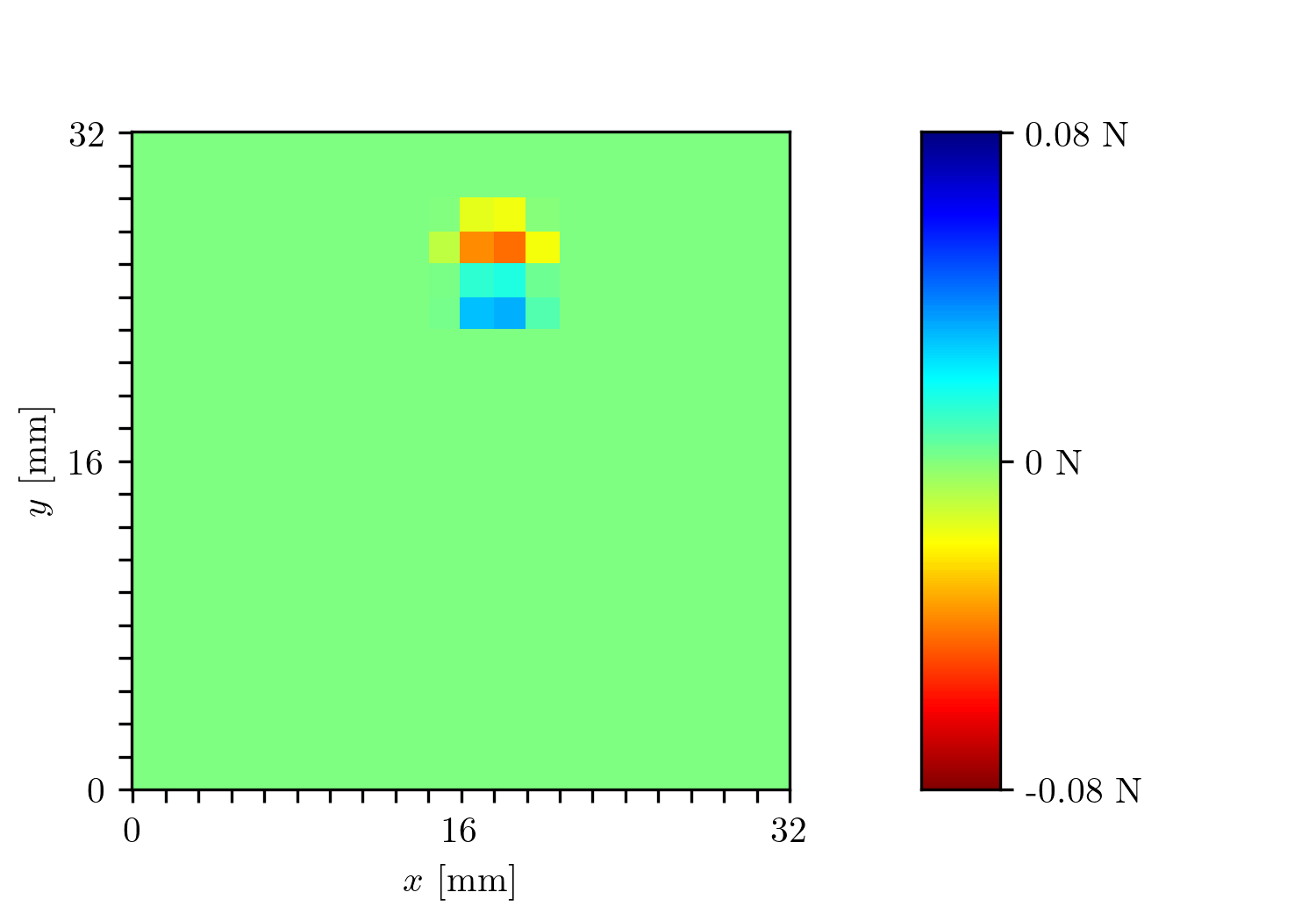} }
	\hfill
	\centering
	\subfloat[$z$ component of the predicted force distribution]{%
	\includegraphics[width=0.5\textwidth]{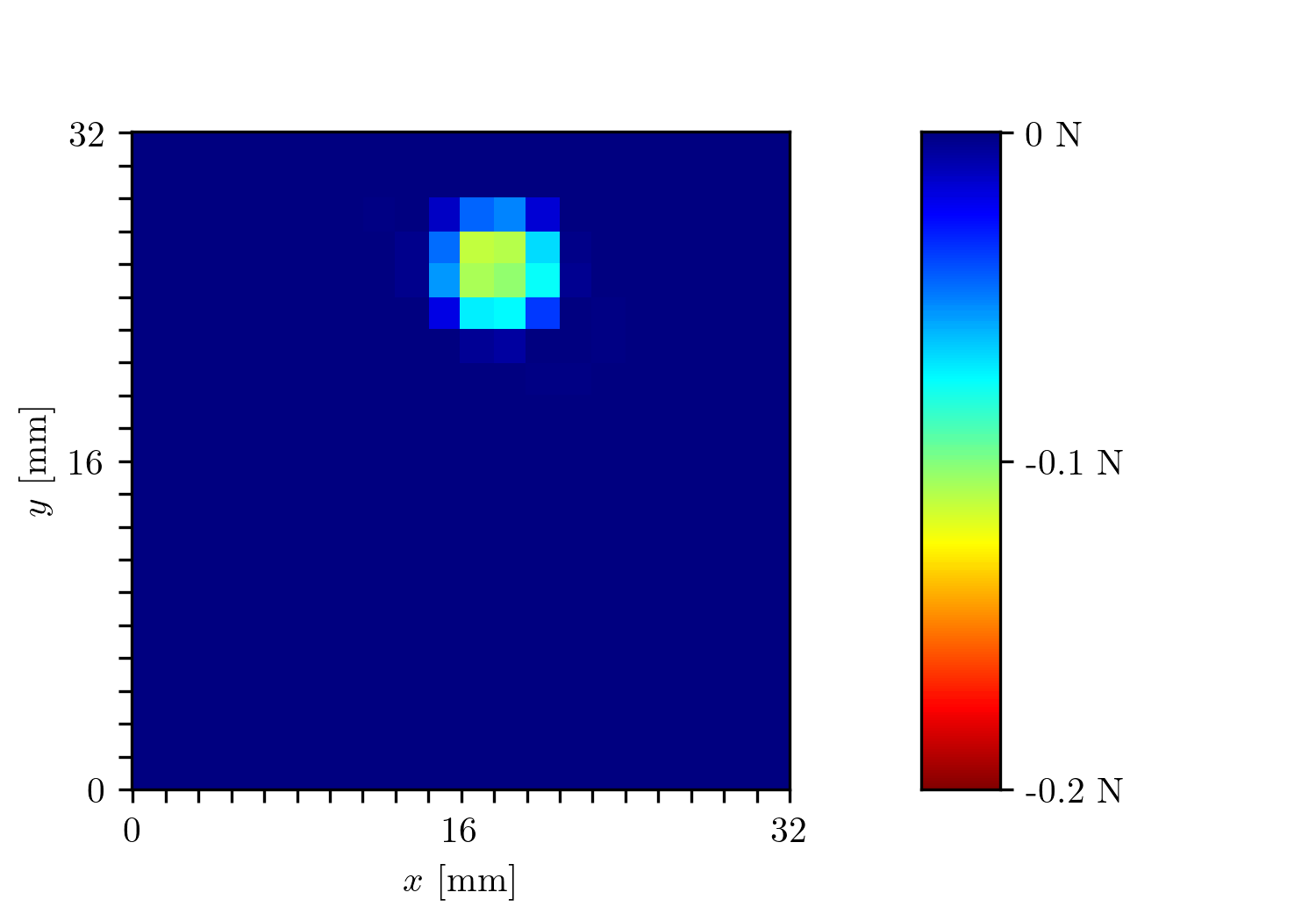} }
	\subfloat[$z$ component of the ground truth force distribution]{%
	\includegraphics[width=0.5\linewidth]{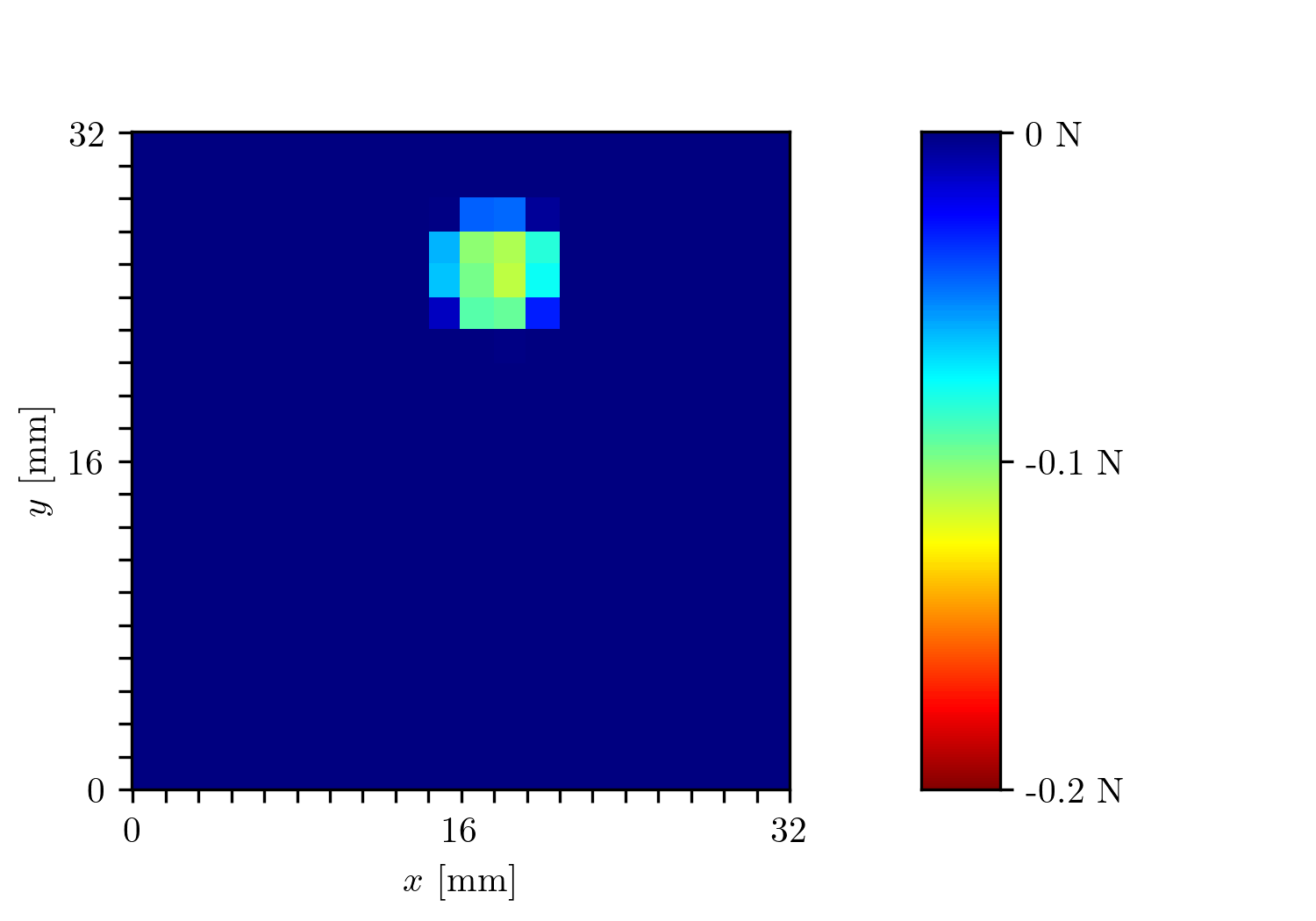} }
	\caption{The plots above show the predicted (left) and ground truth (right) 3D contact force distribution applied to the top surface of the tactile sensor for an indentation in the test set. Note that the axes are defined as in Section \ref{sec:data_collection}, that is, with two perpendicular horizontal axes $x$ and $y$, aligned with two of the top surface edges, and a vertical axis $z$, which is positive pointing from the camera towards the top surface.}
	\label{fig:predictions}
\end{figure*}

\begin{figure*}
	\centering
	\subfloat[$x$ component of the predicted force distribution]{%
		\includegraphics[width=0.5\textwidth]{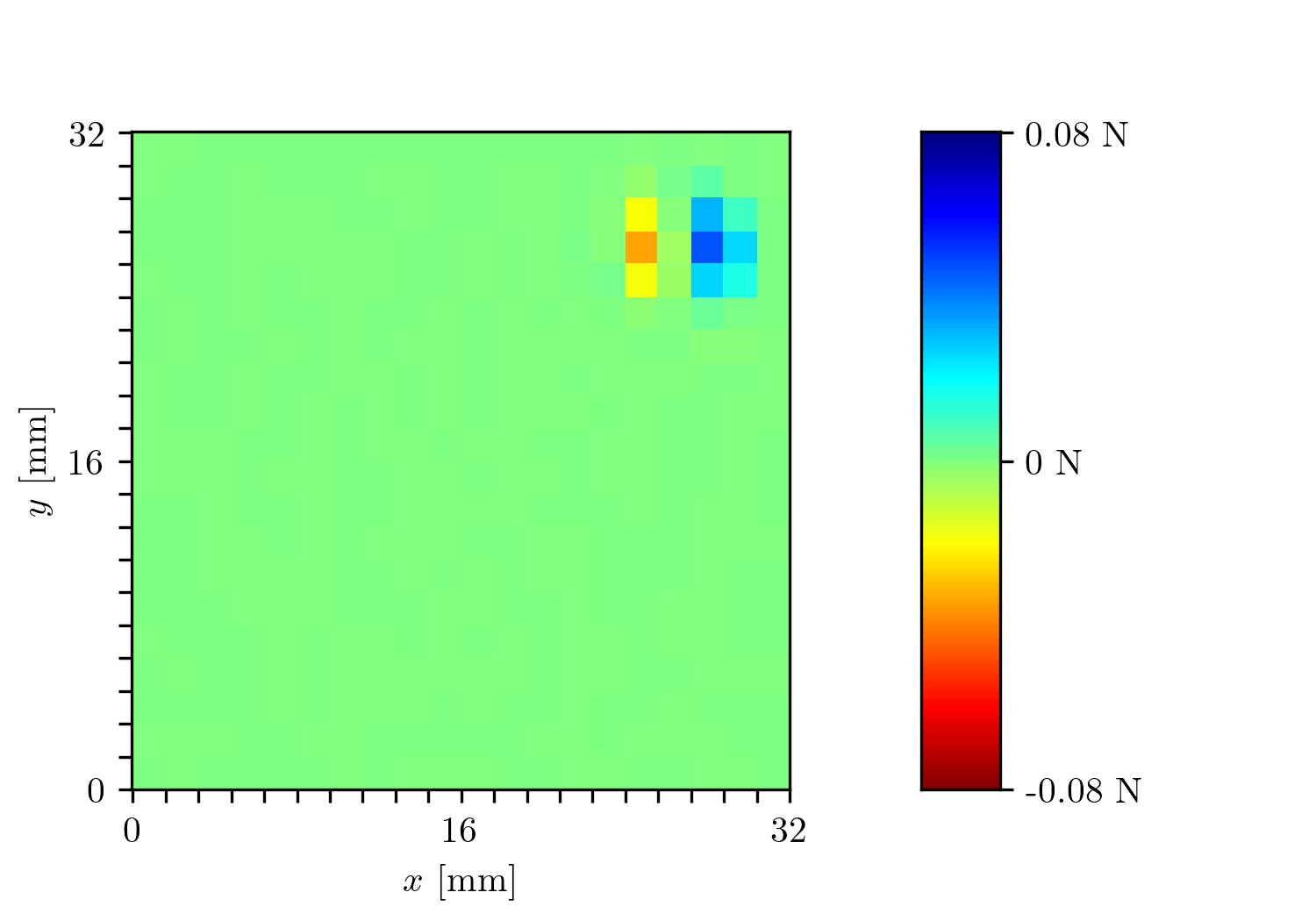} }
	\subfloat[$x$ component of the ground truth force distribution]{%
		\includegraphics[width=0.5\linewidth]{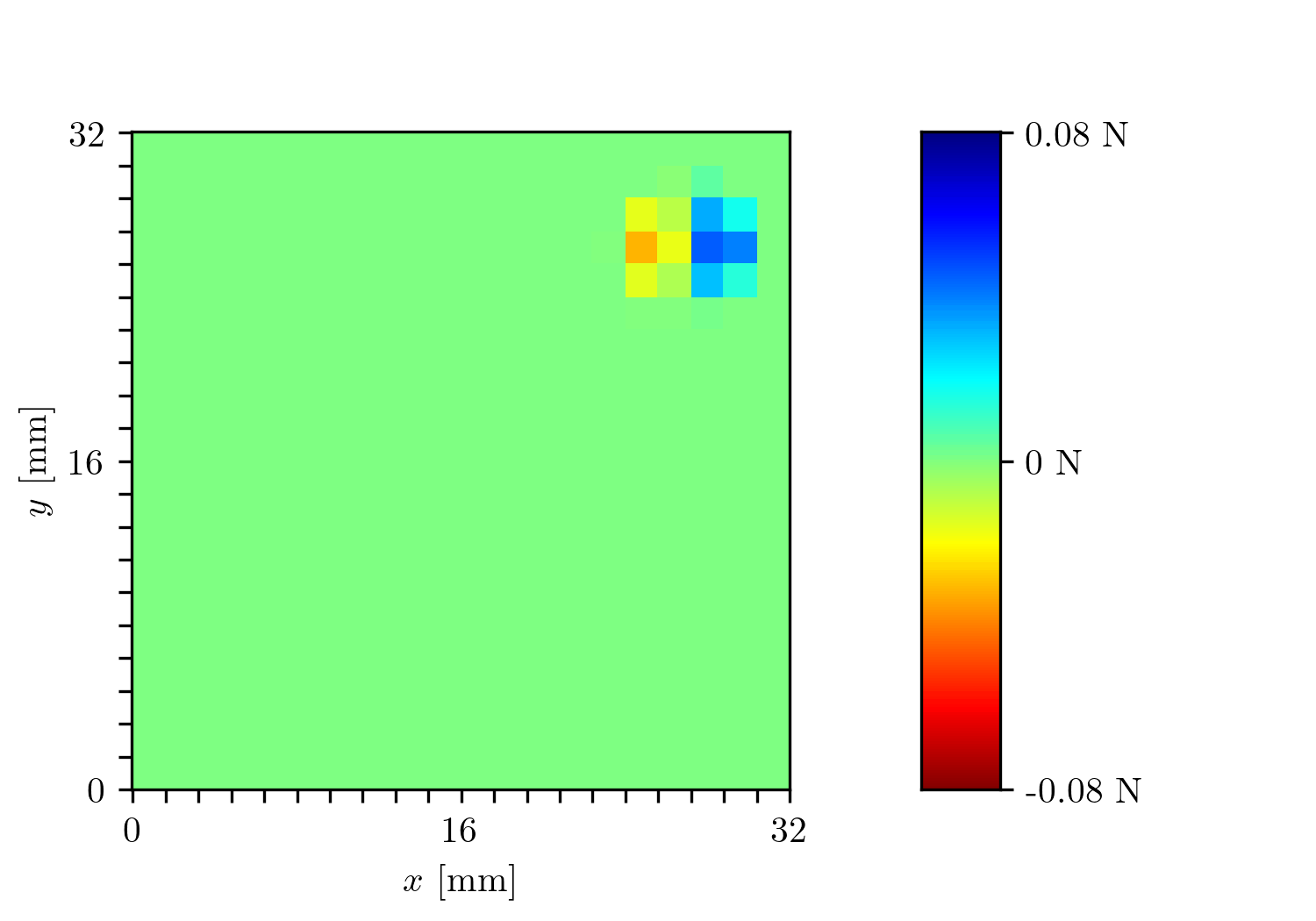} }
	\hfill
	\centering
	\subfloat[$y$ component of the predicted force distribution]{%
		\includegraphics[width=0.5\textwidth]{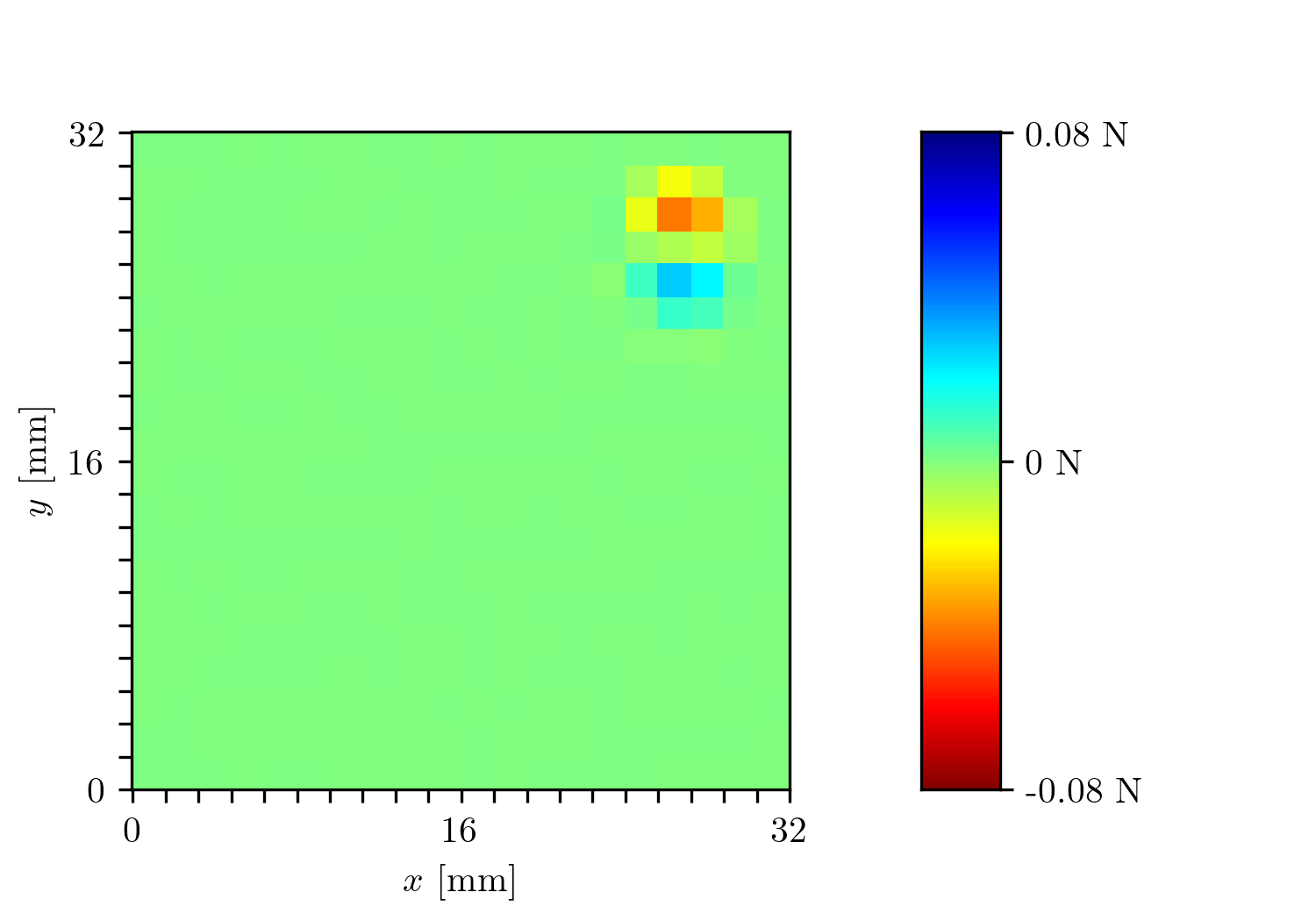} }
	\subfloat[$y$ component of the ground truth force distribution]{%
		\includegraphics[width=0.5\linewidth]{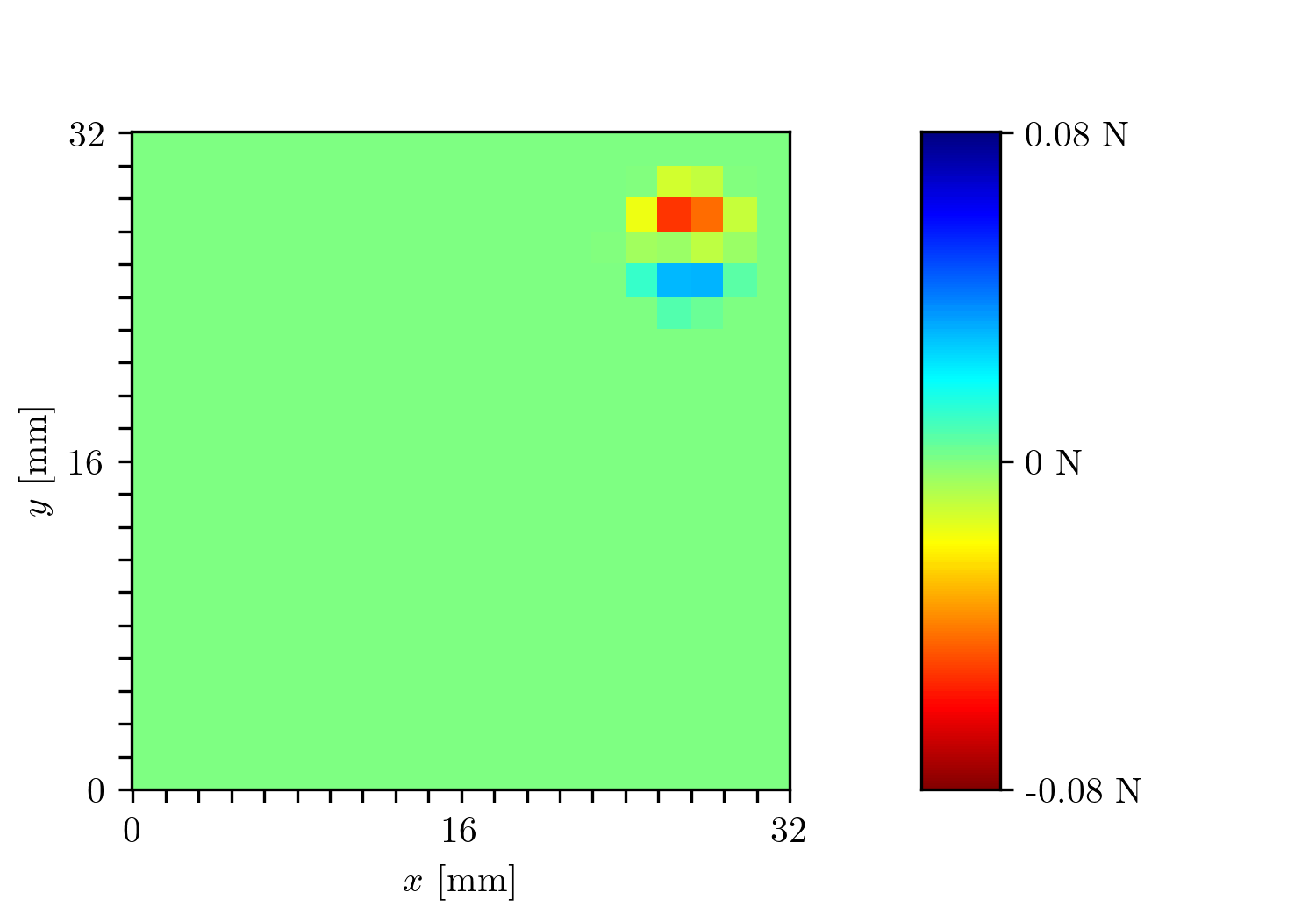} }
	\hfill
	\centering
	\subfloat[$z$ component of the predicted force distribution]{%
		\includegraphics[width=0.5\textwidth]{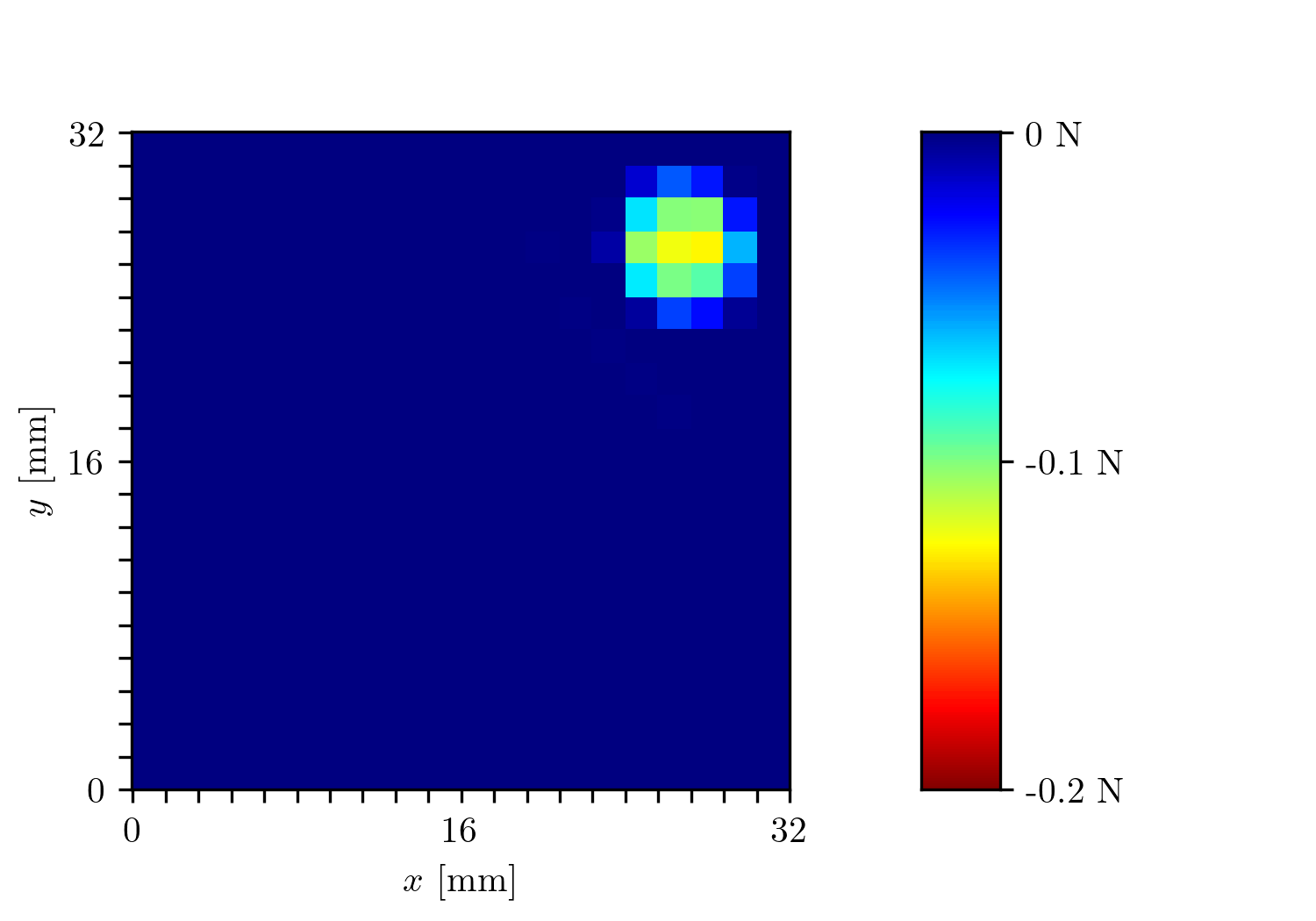} }
	\subfloat[$z$ component of the ground truth force distribution]{%
		\includegraphics[width=0.5\linewidth]{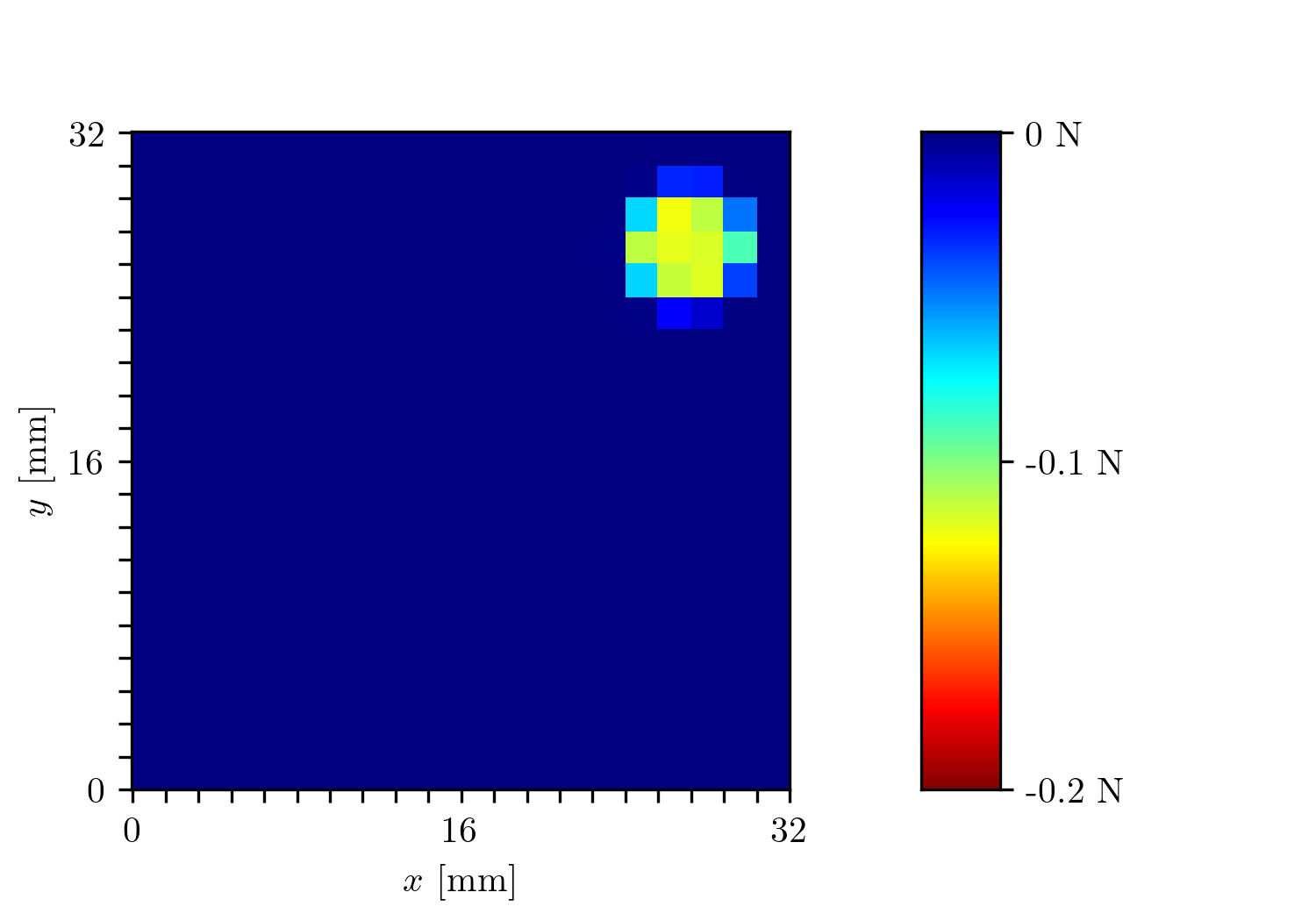} }
	\caption{Similarly to Figure \ref{fig:predictions}, the plots above show the predicted (left) and ground truth (right) 3D contact force distribution applied to the top surface of the tactile sensor for an indentation in the test set. In this case, the indentation was centered at 5 mm from the top and right edges, with the indenter's radius being 5 mm. It can be noted a larger asymmetry in the force distribution, affected by the stiffer edges of the gel.}
	\label{fig:predictions_edge}
\end{figure*}

Moreover, the RMSE on the total force is estimated for both the cases, in which the ground truth is provided either by the FEM simulations ($\text{RMSET}_\text{FEM}$) or the F/T sensor ($\text{RMSET}_\text{F/T}$). The resulting errors from the predictions on the test set are summarized in Table \ref{table:errors} for each axis.
The values in the last row are affected by both the errors introduced by the FEM modeling and the DNN predictions. Note that it is only possible to compute the metrics for the force distribution (first two rows) in relation to the FEM simulations (the F/T sensor only provides total forces, without specific information about the force distribution). As a reference, the ranges of force provided by the ground truth labels are summarized in Table \ref{table:force_ranges}.
Examples of the predicted contact force distribution are shown in Fig. \ref{fig:predictions} and Fig. \ref{fig:predictions_edge}. 

The resulting DNN is deployed on the dual-core laptop computer introduced in Section \ref{sec:introduction}. The entire pipeline yields real-time predictions on an image stream of 40 frames per second, as shown in the experiments available in the video attached to this article. The parallelization of both the optical flow algorithm and the neural network prediction step is not exploited here, but it may be leveraged on commercially available embedded computers provided of GPUs.

\begin{table}[h!]
	\centering
	\caption{Resulting errors on force distribution and total force}
	\label{table:errors}
	\begin{tabular}{c|c|c|c} 
		\hline
		\rule{0pt}{2ex}
		Metric & $x$ & $y$ & $z$\\
		\hline
		\rule{0pt}{3ex}
		RMSE & 0.001 N & 0.001 N & 0.003 N\\ 
		$\text{RMSES}$ & 0.007 N& 0.006 N& 0.016 N \\
		\hline
		\rule{0pt}{2.5ex}
		$\text{RMSET}_\text{FEM}$ & 0.004 N & 0.004 N & 0.045 N\\
		$\text{RMSET}_\text{F/T}$ & 0.025 N & 0.021 N & 0.082 N \\
		\hline
	\end{tabular}
\end{table}

\begin{table}[h!]
	\centering
	\caption{Range of ground truth forces}
	\label{table:force_ranges}
	\begin{tabular}{c|c|c|c} 
		\hline
		\rule{0pt}{2ex}
		Quantity & $x$ & $y$ & $z$\\
		\hline
		\rule{0pt}{3ex}
		Force per bin & [-0.06,0.06] N & [-0.06,0.06] N & [-0.15,0] N\\
		\hline
		\rule{0pt}{2.5ex}
		Total force  & [-0.13,0.13] N & [-0.13,0.13] N & [-1.66,0] N\\
		\hline
	\end{tabular}
\end{table}


\section{Conclusion} \label{sec:conclusion}
This article has presented a strategy to provide ground truth contact force distribution for learning-based tactile sensing. The approach has been evaluated on a vision-based tactile sensor, which is based on tracking particles spread within a soft gel. After the characterization of the hyperelastic materials, which provides accurate material models for FEA, a large number of real indentations and corresponding simulations have been performed to generate a dataset that includes image features and ground truth for the 3D contact force distribution. 
Although the material characterization was performed with considerably different tests and setup (i.e., UA, PS, EB tests) than the indentations considered in the evaluation, the total forces recorded in the experiments are comparable to the ones determined in simulation, showing the generalization potential of the approach proposed.
Note that due to the fact that the simulation labels are assigned to real data obtained from experimental indentations, the experimental setup needs to be carefully arranged. As an example, the alignment of the tactile sensor with the reference axes of the milling machine used for the data collection is crucial for obtaining good performance.

As shown in Section \ref{sec:learning}, the dataset generated with the strategy proposed in this article can be used to train a DNN for accurate reconstruction of the force distribution applied to the surface of the tactile sensor. Although in these experiments the DNN has been trained and evaluated on a sample indenter, the techniques presented here are directly applicable to generic shapes and indentations, including multiple and distinct contact areas. However, this would likely require the collection of a dataset under various contact conditions, involving interactions with complex objects. Therefore, the generalization capabilities of this strategy will be object of future work.

\appendices
\section{Strain-rate and shelf-time dependent properties of Ecoflex GEL}\label{sec:appendix}
\begin{figure}[h!]
	\subfloat[Strain-rate dependence in uniaxial tension for the Ecoflex GEL]{%
	\includegraphics[scale=1.0]{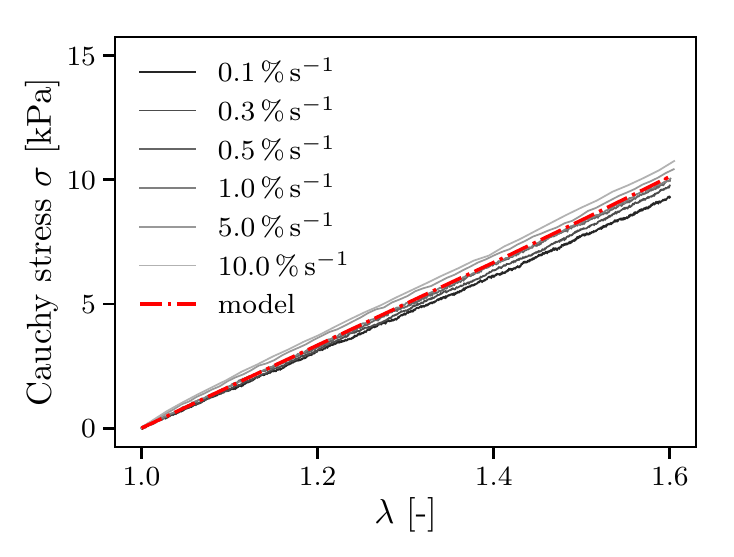}}
	\hfill
	\subfloat[Age-dependent mechanical properties of the Ecoflex GEL]{%
	\includegraphics[scale=1.0]{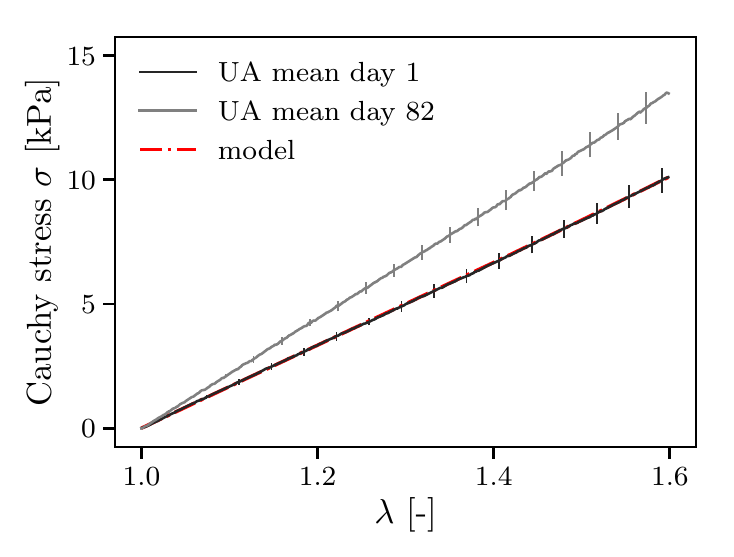}}
	\caption{Uniaxial tension (UA) tests showing (a) strain-rate and (b) shelf-time (aging) dependence on the mechanical properties of Ecoflex GEL.}
	\label{fig:rate_aging_ecoflex}
\end{figure}

The additional experimental data on the rate-dependence and the aging effect on the mechanical behavior of the Ecoflex GEL are shown in Fig. \ref{fig:rate_aging_ecoflex}. While the strain-rate dependence remains relatively low over the three decades analyzed (Fig. \ref{fig:rate_aging_ecoflex}a), a significant stiffening after 9 weeks of storage at room temperature is evident due to material aging (Fig. \ref{fig:rate_aging_ecoflex}b). Longer curing times and higher curing temperatures may be used to approach the final, curing-independent material properties \cite{placet_pdms_aging, hopf_pdms_characterization}.

\section*{Acknowledgment}
The authors would like to thank Michael Egli for the manufacturing support and Francesco Filotto for his insights about the generation of the ground truth labels.

\bibliographystyle{ieeetran}
\bibliography{references}

\begin{IEEEbiography}[{\includegraphics[width=1in,height=1.25in,clip,keepaspectratio]{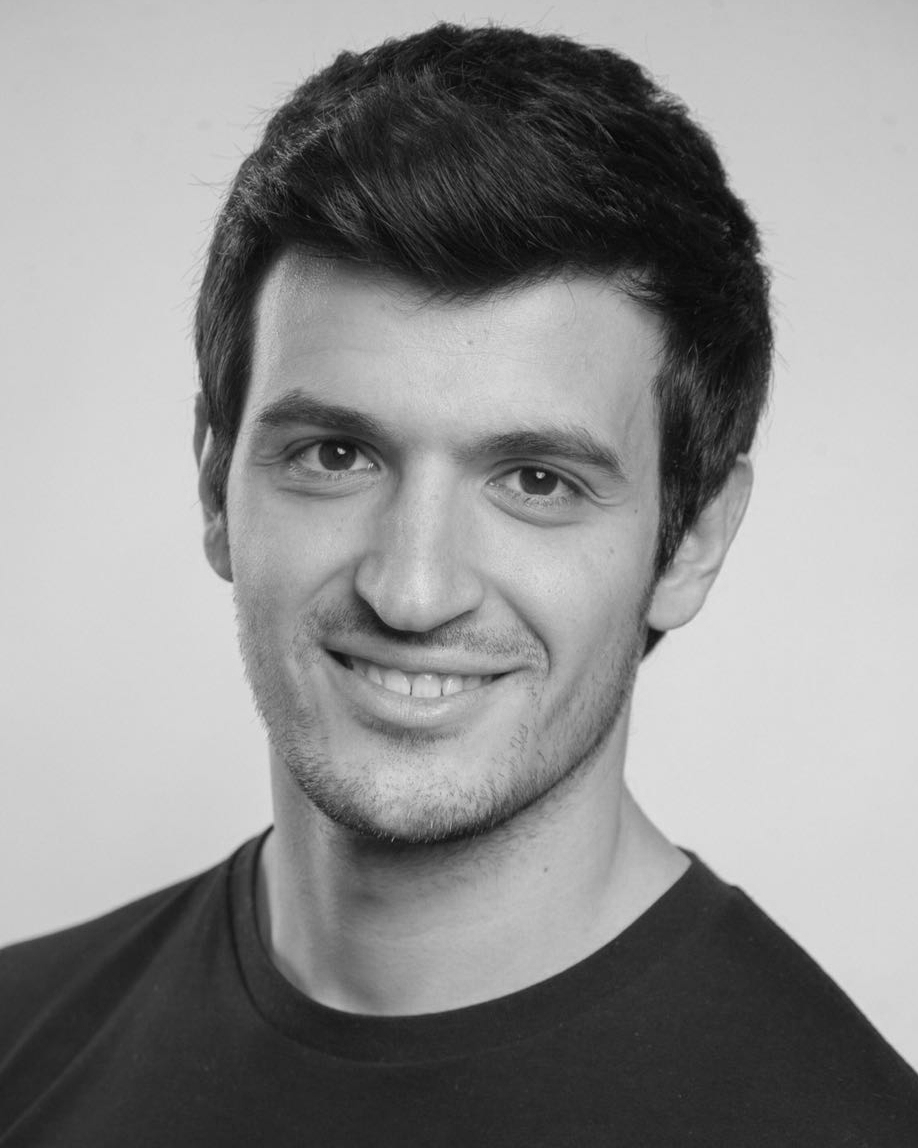}}]{Carmelo Sferrazza} (S'17) received the B.Sc. degree in Automation Engineering from Politecnico di Milano, Milan, Italy, in 2014, and the M.Sc. degree in Robotics, Systems and Control from ETH Zurich, Zurich, Switzerland, in 2016. During his studies, he completed a double-degree program and received the B.Eng. in Electronics and Information Engineering from the Tongji University, Shanghai, China. 
	
He is currently pursuing a doctorate degree in the Institute for Dynamic Systems and Control at ETH Zurich. His current research interests include the design and the development of vision-based, data-driven tactile sensors, and learning-based model predictive control. 

Mr. Sferrazza received the ETEL Award for his master's thesis on model predictive control of an unmanned aerial vehicle.
\end{IEEEbiography}

\begin{IEEEbiography}[{\includegraphics[width=1in,height=1.25in,clip,keepaspectratio]{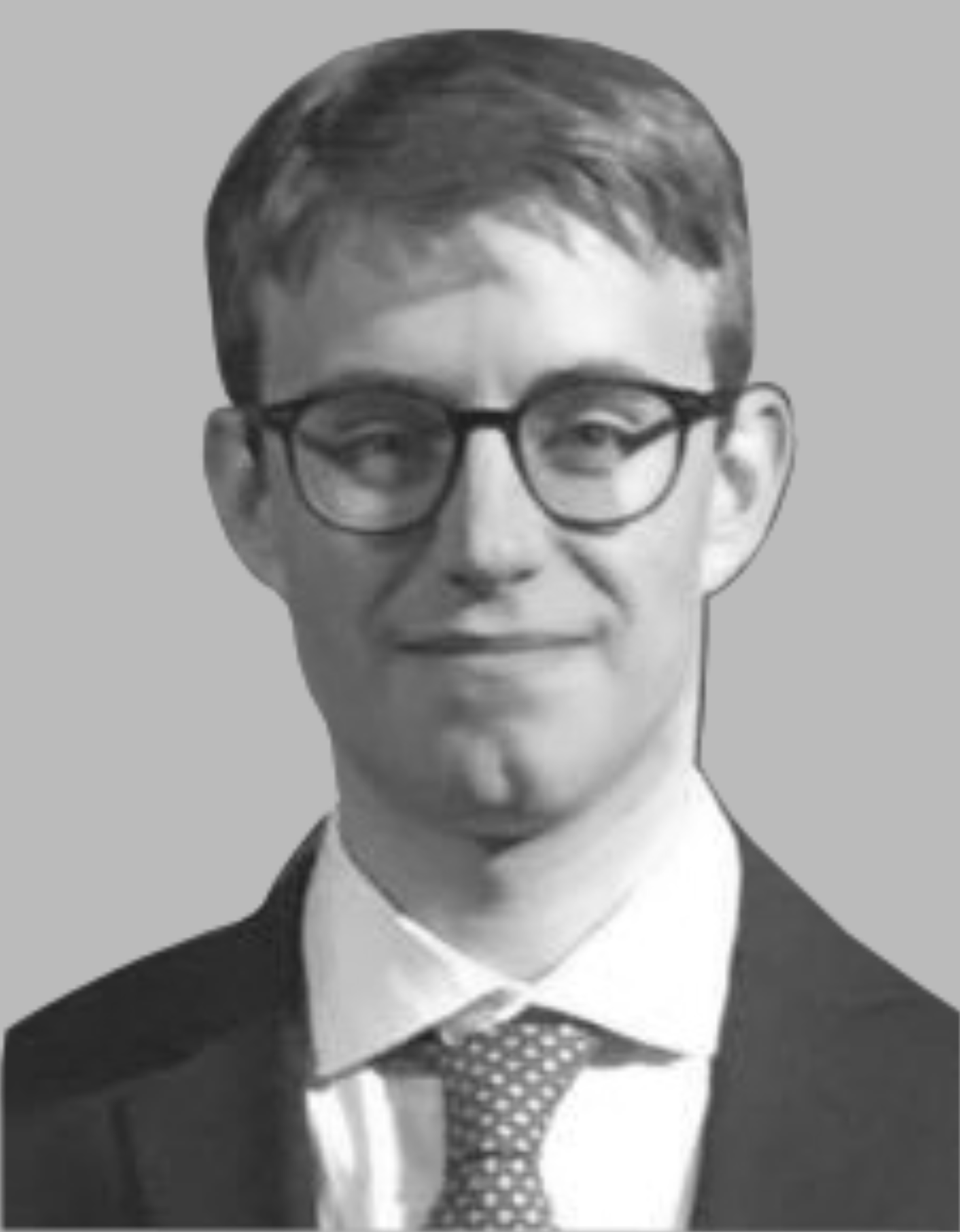}}]{Adam Wahlsten} received his M.Sc. degree in engineering physics from Lund University, Lund, Sweden, in 2017, focusing on computational solid and fluid mechanics. During his studies, he completed an exchange at ETH Zurich, Zurich, Switzerland, where he also wrote his master's thesis on multiphase modeling of soft tissues.
	
He is currently working toward the doctorate degree at the Institute for Mechanical Systems at ETH Zurich. His research focuses on the mechanical characterization and biomechanical modeling of the skin and skin-like tissue-engineered materials.
\end{IEEEbiography}

\begin{IEEEbiography}[{\includegraphics[width=1in,height=1.25in,clip,keepaspectratio]{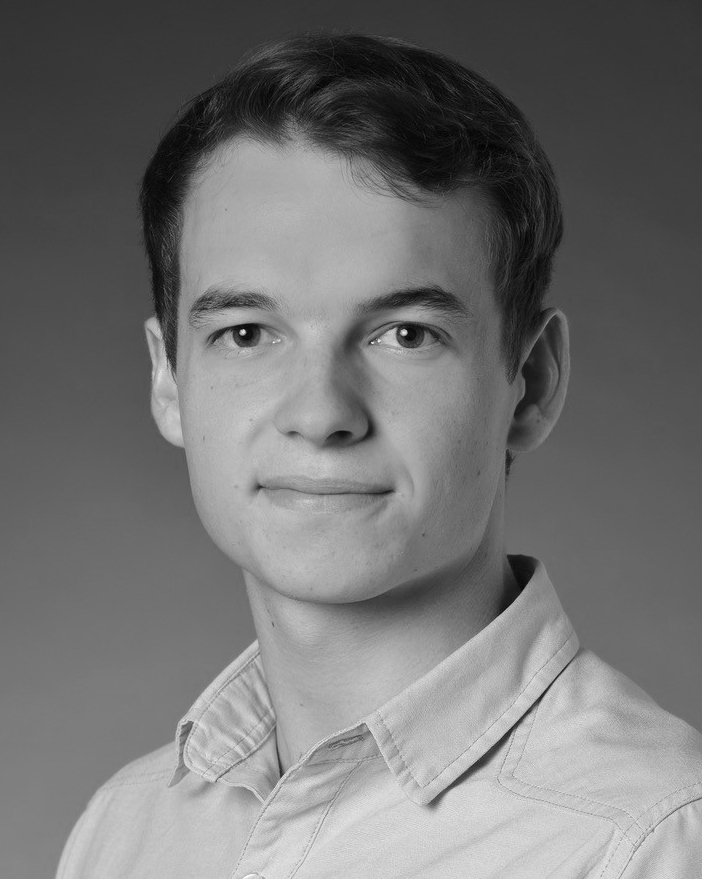}}]{Camill Trueeb} received his B.Sc. degree in mechanical engineering from ETH Zurich, Zurich, Switzerland, in 2017, where he is currently pursuing his M.Sc. degree, focusing on the fields of robotics, systems and control. During his studies, he worked as a teaching assistant in control systems and participated in a project to develop a stretchable volume sensor for the human bladder. 
	
After doing a semester project on ground truth acquisition for vision-based tactile sensors, he is currently writing his master's thesis on deep learning calibration approaches for tactile sensors.
\end{IEEEbiography}

\begin{IEEEbiography}[{\includegraphics[width=1in,height=1.25in,clip,keepaspectratio]{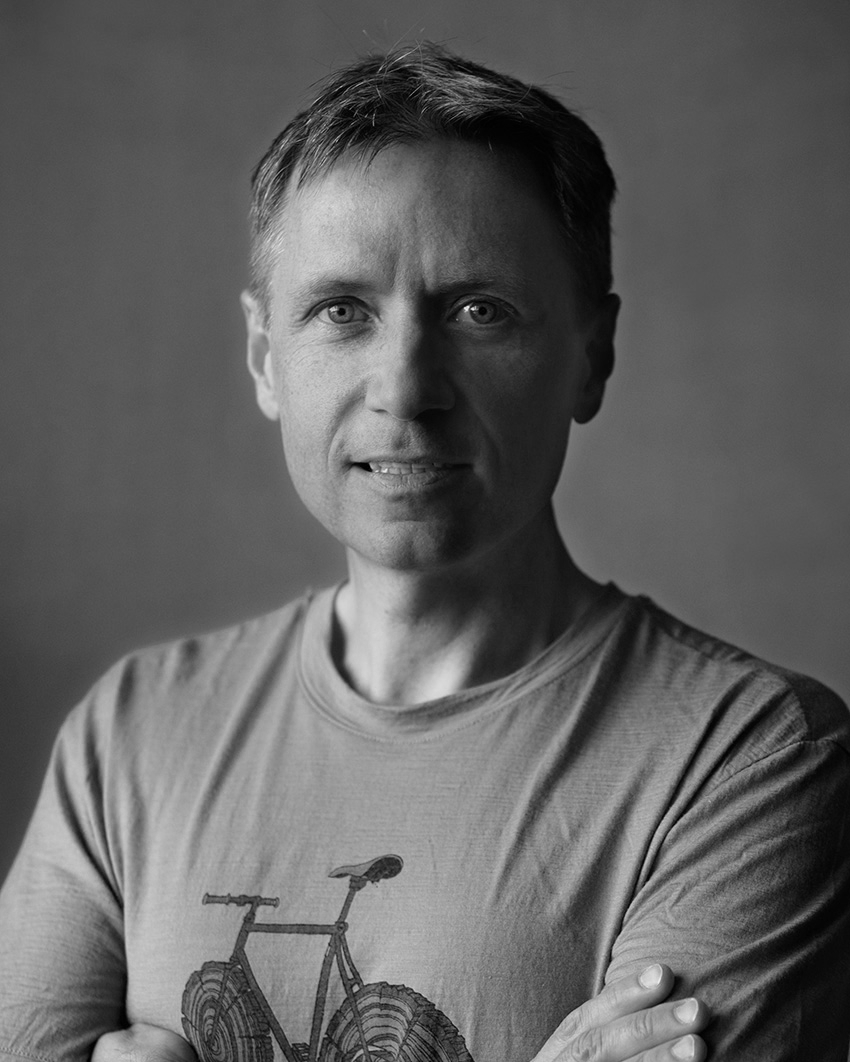}}]{Raffaello D'Andrea} (F'10) received the B.Sc. degree in Engineering Science from the
	University of Toronto in 1991, and the M.S. and Ph.D. degrees in Electrical
	Engineering from the California Institute of Technology in 1992 and 1997. He
	was an assistant, and then an associate, professor at Cornell University from
	1997 to 2007. While on leave from Cornell, from 2003 to 2007, he co-founded
	Kiva Systems (now Amazon Robotics), where he led the systems architecture, 
	robot design, robot navigation and coordination, and control algorithms development. 
	He is currently professor of Dynamic Systems and Control at ETH Zurich, 
	and chairman of the board at Verity Studios AG.
\end{IEEEbiography}

\EOD

\end{document}